\documentclass[sigconf]{acmart}

\usepackage{bm}
\usepackage[linesnumbered,boxed,vlined,ruled,noend]{algorithm2e}
\usepackage{multirow}
\usepackage{balance}
\usepackage{tikz}
\usepackage{url}
\usepackage{xspace}
\newcommand{\sys}{\textsc{DGMLP}\xspace}
\newcommand*{\circled}[1]{\lower.7ex\hbox{\tikz\draw (0pt, 0pt)circle (.5em) node {\makebox[1em][c]{\small #1}};}}
\usepackage{xspace}
\usepackage{multirow}
\usepackage{bm}
\usepackage{bbm}
\usepackage{color, soul}
\usepackage{subfigure}
\usepackage{graphicx}
\usepackage{listings}
\usepackage{xcolor}
\usepackage{makecell}
\usepackage{color, soul}

\usepackage{bm}
\usepackage{booktabs}
\usepackage{wrapfig}
\usepackage{enumitem}
\usepackage{CJKutf8}
\usepackage{textcomp}
\usepackage{hyperref}

\newcommand{\para}[1]{{\vspace{2pt} \bf \noindent #1 \hspace{1pt}}}

\usepackage[linesnumbered,vlined,ruled,noend]{algorithm2e}

\AtBeginDocument{%
  \providecommand\BibTeX{{%
    \normalfont B\kern-0.5em{\scshape i\kern-0.25em b}\kern-0.8em\TeX}}}

\begin{document}
\title{Evaluating Deep Graph Neural Networks}


\renewcommand{\shorttitle}{\sys}
\author{Wentao Zhang$^\dagger$, Zeang Sheng$^{\dagger}$, Yuezihan Jiang$^{\dagger}$, Yikuan Xia$^{\dagger}$, Jun Gao$^{\dagger}$,
Zhi Yang$^{\dagger \S}$, 
Bin Cui$^{\dagger \S}$}
\affiliation{
 {{$^\dagger$}{School of EECS \& Key Laboratory of High Confidence Software Technologies, Peking University}}~~~~~\\
 {{$^{\S}$}{Center for Data Science, Peking University \& National Engineering Laboratory for Big Data Analysis and Applications}}}
\affiliation{
$^\dagger$\{wentao.zhang, shengzeang18, jiangyuezihan, 2101111522, gaojun, yangzhi, bin.cui\}@pku.edu.cn
}
\renewcommand{\shortauthors}{Zhang et al.}


\begin{abstract}
Graph Neural Networks (GNNs) have already been widely applied in various graph mining tasks. 
However, they suffer from the shallow architecture issue, which is the key impediment that hinders the  model performance improvement.
Although several relevant approaches have been proposed, none
of the existing studies provides an in-depth understanding of the root causes of performance degradation in deep GNNs.
In this paper, we conduct the first systematic experimental evaluation to present the fundamental limitations of shallow architectures. Based on the
experimental results, we answer the following
two essential questions: (1) what actually leads to the compromised performance of deep GNNs; (2) when we need and how to build deep GNNs. The answers to the above questions provide empirical insights and guidelines for researchers to design deep and well-performed GNNs.
To show the effectiveness of our proposed guidelines, we present Deep Graph Multi-Layer Perceptron (DGMLP), a powerful approach (a paradigm
in its own right) that helps guide deep GNN designs.
Experimental results demonstrate three advantages of \sys: 1) high accuracy -- it achieves state-of-the-art node classification performance on various datasets; 2) high flexibility -- it can flexibly choose different propagation and transformation depths according to graph size and sparsity; 3) high scalability and efficiency -- it supports fast training on large-scale graphs.
Our code is available in \href{https://github.com/zwt233/DGMLP}{https://github.com/zwt233/DGMLP}.

\end{abstract}



\settopmatter{printacmref=false, printfolios=false}
\maketitle
{\fontsize{8pt}{8pt} \selectfont
\textbf{ACM Reference Format:}\\
Wentao Zhang, Zeang Sheng, Yuezihan Jiang, Yikuan Xia, Jun Gao, Zhi Yang, Bin Cui. 2021. Evaluating Deep Graph Neural Networks. In \textit{Proceedings of  ACM Conference (Conference ’21). } ACM, New York, NY, USA, 14 pages. https://doi.org/10.1145/xxxxxxx.xxxxxxx }


\section{Introduction}

The recent success of Graph Neural Networks (GNNs)~\cite{zhang2020deep} has boosted researches on knowledge discovery and data mining on graph data. 
Designed for graph-structured data, GNNs provide a universal way to tackle node-level, edge-level, and graph-level tasks~\cite{zhou2020graph}, including social analysis~\cite{qiu2018deepinf, fan2019graph, huang2021knowledge}, chemistry and biology~\cite{DBLP:conf/nips/DaiLCDS19, DBLP:conf/iclr/BradshawKPSH19, DBLP:conf/kdd/Do0V19}, recommendation~\cite{monti2017geometric, wu2020graph, yin2019deeper}, natural language processing~\cite{bastings2017graph, wu2021graph, vashishth2020graph}, and computer vision~\cite{qi2018learning, shi2019skeleton, sarlin2020superglue}. 

\para{Graph Convolution.} 
The key to the success of most GNNs lies in the convolution operations, which propagate the neighbor information to a target node in an iterative manner~\cite{li2018deeper}. 
This convolution operation can be further decomposed into two sequential steps: \textit{embedding propagation} (EP) and \textit{embedding transformation} (ET).
The EP step can be viewed as a special form of Laplacian smoothing~\cite{DBLP:journals/corr/abs-1905-09550, DBLP:conf/ijcai/ZhouSHZZH20}, which combines the embeddings of a node itself and its one-hop neighbors.
The embeddings of nodes within the same connected component would become similar after applying the smoothing operation, which greatly eases the downstream tasks and it is the crucial reason why GNNs work well. 
The ET step applies neural networks as a powerful feature extractor.
Taking the widely-used Graph Convolutional Network (GCN)~\cite{kipf2016semi} as an example, through stacking $k$ convolution operations (i.e., layers), each labeled node in GCN can utilize the information from its $k$-hop neighborhood, and thus improve the model performance by getting more unlabeled nodes involved in the training process.

Despite the remarkable success, GNNs suffer from a fundamental limitation that they tend to perform worse as they become deeper with
more convolution operations. 
As a result, recent GNNs usually have shallow architectures (e.g., 2 or 3 stacked convolution operations), which ignore the distant connections and can only exploit local structural information.
Moreover, under the semi-supervised setting where few labels are given, a shallow GNN cannot effectively propagate the labels to the entire graph, which limits the full potentials of EP and ET operations on graphs.


\para{Misconceptions.} 
Existing researches usually attribute the above limitation to the following reasons:

(1) \textit{Over-smoothing.} 
Notice that the EP operation makes
the features of nodes in the same connected component similar. If a GNN is deep with
many convolution operations, the output features may be over-smoothed, and nodes from different clusters may become
indistinguishable.
Most existing works~\cite{feng2020graph,chen2020measuring,DBLP:conf/iclr/ZhaoA20, godwin2021very, rong2019dropedge, zeng2020deep, min2020scattering, DBLP:conf/icml/0001RGBWR21, DBLP:conf/iclr/ChienP0M21, zhou2020towards,hou2019measuring, DBLP:conf/icml/BeainiPLHCL21, yan2021two, cai2020note} consider over-smoothing as the main cause of performance degradation of deep GNNs. 

(2) \textit{Over-fitting (or gradient vanishing).} A few methods~\cite{rong2019dropedge,li2019deepgcns,zhou2020understanding,yang2020revisiting} argue that the main difficulty in constructing deep GNNs lies in over-fitting or gradient vanishing, especially for small datasets.

(3) \textit{Entanglement.} Several recent studies~\cite{liu2020towards} 
state that the key factor compromising the performance of deep GNNs is the entanglement of EP and ET inside each GNN layer. Many disentangled GNN variants, such as APPNP ~\cite{klicpera2018predict}, DAGNN~\cite{liu2020towards}, GBP~\cite{chen2020scalable}, AP-GCN~\cite{spinelli2020adaptive}, Grand~\cite{feng2020graph}, and $\text{S}^2$GC~\cite{zhu2021simple}, disentangle EP and ET and obtain better performance by only increasing the number of EP operations. However, the insights and explanations of why the entanglement might hinder the construction of deep convolution have not been well understood.


\para{Questions Investigated.} In this experimental and analysis paper, we thoroughly understand
and to some extent address the fundamental limitation
of deep convolution on graph data, in order
to lay a solid ground for future research developments in the
field. Specifically, we investigate the three key research questions:
\begin{itemize}
\item[\textbf{RQ1}:] What actually limits the stacking of deep convolution operations in GNN designs, and why does disentangling EP and ET could allow more EP operations?



\item[\textbf{RQ2}:] What are the insights and guidelines for researchers to design deep GNNs? 
In particular, when and how to enlarge the number of EP and ET operations?


\item[\textbf{RQ3}:] With the help of the proposed insights and guidelines, can we stack more EP and ET operations and outperform the state-of-the-art GNNs?

\end{itemize}

\para{Contributions.} To answer the above research questions, we first conduct a comprehensive evaluation to revise the common misconceptions and identify the root causes of the shallow architecture issue of most existing GNNs. Based on the above analysis, we further obtain helpful insights and guidelines to construct deep GNNs. Our main findings and contributions are summarized as follows.
\begin{itemize}
\item[\textbf{C1}:] We clarify the concept of depth (i.e., layer) by separating and considering the two different depths of deep GNN design: \textit{the number of stacked EP operations and ET operations}. Through experimental evaluations, we find that a large number of EP operations leads to \textit{over-smoothing} whereas a large number of ET operations leads to \textit{model degradation} in the current GNN models. Moreover, the latter is more sensitive than the former.
Therefore, in the case where EP and ET are entangled (e.g., GCN and GraphSAGE~\cite{hamilton2017inductive}), we find that the failures of deep GNNs are mainly caused by \textit{model degradation} rather than \textit{over-smoothing}.
When EP and ET are disentangled, one can individually increase the number of EP operations while avoiding the influence of \textit{model degradation} introduced by a large number of ET operations. 

\item[\textbf{C2}:] Our experimental evaluation implies that we need a large number of EP operations when the graph is sparse on some properties like edge, label, and feature. Meanwhile, we need to enlarge the number of ET operations to improve the feature extraction ability when the graph is large.
To this end, to stack more EP operations, we propose a node-adaptive combination mechanism of propagated features under EP operations of different steps.
To design models with a large number of ET operations, we introduce residual connections into ET operations to alleviate the \textit{model degradation} issue.

\item[\textbf{C3}:] We propose \sys, a novel deep convolution method that adopts the composition of a node-adaptive combination mechanism and residual connections to stack a large number of both EP and ET operations based on our findings on deep convolution. 
We validate the effectiveness of \sys on seven public datasets. Experimental results demonstrate that \sys outperforms the state-of-the-art GNNs while maintains high scalability and efficiency. 
\end{itemize}
To the best of our knowledge, this paper is the first experimental evaluation work that identifies the root causes of shallow GNN architectures.
Our findings and implications open up a new perspective on designing deep GNNs for graph-structured data.



\section{Preliminary}

In this section, we firstly introduce the notations and the problem formulation in this paper. Then we explain the correlation between GCN and MLP, and at the same time, review current GNNs from the pattern of EP and ET operations.

\subsection{Problem Formalization}
Throughout this paper, we consider an undirected graph $\mathcal{G}$ = ($\mathcal{V}$,$\mathcal{E}$) with $|\mathcal{V}| = N$ nodes and $|\mathcal{E}| = M$ edges. We denote by $\mathbf{A}$ the adjacency matrix of $\mathcal{G}$, weighted or not. Each node can possibly have a feature vector of size $d$, stacked up to an $N \times d$ matrix $\mathbf{X}$. $\mathbf{D}=\operatorname{diag}\left(d_{1}, d_{2}, \cdots, d_{n}\right) \in \mathbb{R}^{N \times N}$ denotes the degree matrix of $\mathbf{A}$, where $d_{i}=\sum_{j \in \mathcal{V}} \mathbf{A}_{i j}$ is the degree of node $i$. In this paper, we focus on the node classification task under semi-supervised setting. 
Suppose $\mathcal{V}_l$ is the labeled set, the goal is to predict the labels for nodes in the unlabeled set $\mathcal{V}_u$ under the label supervision of $\mathcal{V}_l$.

\subsection{Convolution on Graphs}

\para{Convolution.}
Images can be considered as a specific type of graphs where pixels are connected by adjacent pixels, and the neighbors of each pixel are determined by the filter size. 
The 2D convolution in convolutional neural networks (CNN) takes the weighted average pixel values of each pixel along with its neighbors.
Similarly, one may perform graph convolutions by the weighted averaging of a node’s neighborhood information~\cite{wu2020sur}.

Recent works on GNNs, especially graph convolutional networks, have attracted considerable attention for their remarkable success in graph representation learning. For example, RecGNNs~\cite{li2015gated} and ConvGNNs~\cite{kipf2016semi} are able to extract high-level node embeddings by propagation. 
Based on the intuitive assumption that locally connected nodes are likely to have the same label,
GNN iteratively propagates the information of each node to its adjacent nodes.
For example, each graph convolution operation in GCN firstly propagates the node embeddings to their neighborhoods and then transforms their propagated node embeddings:
\begin{equation}
\small
    \mathbf{X}^{(k+1)}=\sigma\big(\mathbf{\hat{A}}\mathbf{X}^{(k)}\mathbf{W}^{(k)}\big), \qquad \mathbf{\hat{A}} = \widetilde{\mathbf{D}}^{\frac{1}{2}}\widetilde{\mathbf{A}}\widetilde{\mathbf{D}}^{-\frac{1}{2}},
    \label{eq_GC}
\end{equation}

\noindent where $\mathbf{X}^{(k)}$ and $\mathbf{X}^{(k+1)}$ are the node embeddings of layer $k$ and $k+1$, respectively.
$\widetilde{\mathbf{A}}=\mathbf{A}+\mathbf{I}_{N}$ is the adjacency matrix of the undirected graph $\mathcal{G}$ with added self-connections, where $\mathbf{I}_{N}$ is the identity matrix.
And $\mathbf{\hat{A}}$ is the normalized adjacency matrix used to propagate information, and $\mathbf{\widetilde{D}}$ is the corresponding degree matrix.

By setting different $r$ in $\mathbf{\hat{A}} = \widetilde{\mathbf{D}}^{r-1}\widetilde{\mathbf{A}}\widetilde{\mathbf{D}}^{-r}$, we can employ different normalized strategies accordingly, such as the symmetric normalization $\widetilde{\mathbf{D}}^{\frac{1}{2}}\widetilde{\mathbf{A}}\widetilde{\mathbf{D}}^{-\frac{1}{2}}$~\cite{kipf2016semi}, the transition probability $\widetilde{\mathbf{A}}\widetilde{\mathbf{D}}^{-1}$~\cite{DBLP:conf/iclr/ZengZSKP20}, and the reverse transition probability $\widetilde{\mathbf{D}}^{-1}\widetilde{\mathbf{A}}$~\cite{xu2018representation}. We adopt $\mathbf{\hat{A}}  = \widetilde{\mathbf{D}}^{-1/2}\widetilde{\mathbf{A}}\widetilde{\mathbf{D}}^{-1/2}$ in this work.

\begin{figure}[tpb!]
	\centering
	\includegraphics[width=.95\linewidth]{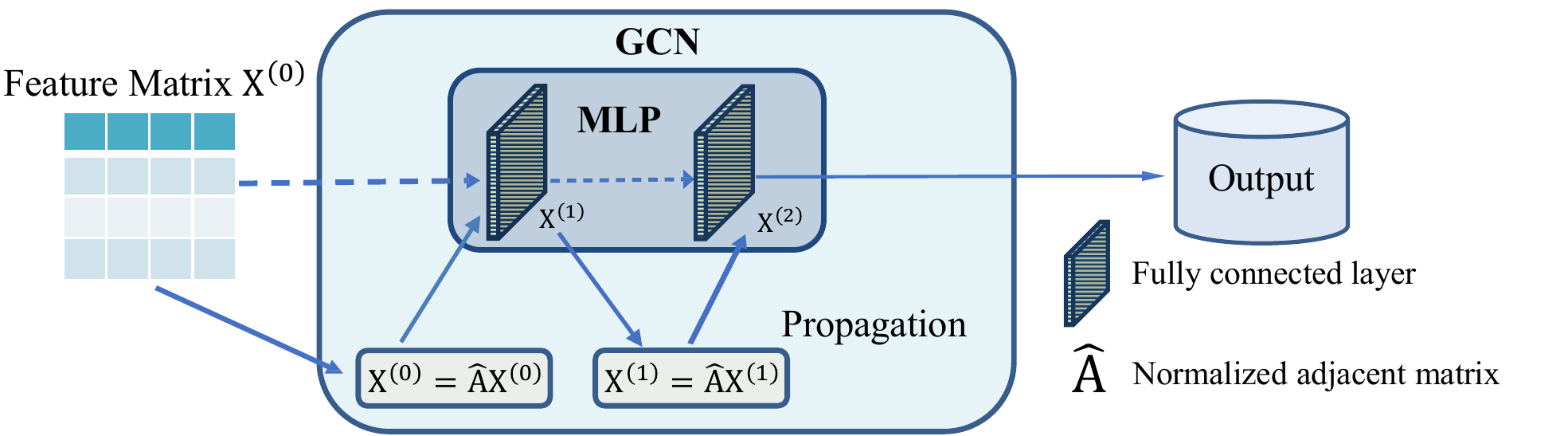}
	\caption{\small  The relationship between GCN and MLP.}
	\label{fig:GCN}
\end{figure}

\para{EP and ET Operations.}
Each graph convolutional layer in GNN can be divided into two operations: Embedding Propagation (EP) and Embedding Transformation (ET), which correspondingly results in two GNN depths: $D_p$ and $D_t$.
Concretely, GNN first executes the embedding propagation with the normalized adjacency matrix $\mathbf{\hat{A}}$ to the embedding matrix $\mathbf{X}$:
\begin{equation}
\small
    \text{EP}(\mathbf{X})= \mathbf{\hat{A}}\mathbf{X},
    \label{eq_EP}
\end{equation}

Then, the propagated feature $\hat{\mathbf{X}} = \text{EP}(\mathbf{X})$ will be transformed with the learnable transformation matrix $\mathbf{W}$ and the activation function $\sigma(\cdot)$:
\begin{equation}
\small
\begin{aligned}
    \text{ET}(\hat{\mathbf{X}})= \sigma(\hat{\mathbf{X}}\mathbf{W}),
    \label{eq_ET}
    \end{aligned}
\end{equation}

Fig.~\ref{fig:GCN} shows the framework of a two-layer GCN. GCN will degenerate to MLP if $\mathbf{\hat{A}}$ is the identity matrix, which is equal to removing the EP operation in all GCN layers.

\subsection{Convolution Pattern}
According to whether the model disentangles the EP and ET operations, current GNNs can be classified into two major categories: entangled and disentangled, and each can be further classified into two smaller categories based on the order of EP and ET operations:

\para{Entangled Graph Convolution.}
Entangled Propagation and Transformation (\textbf{EPT}) pattern is widely adopted by mainstream GNNs, like GCN~\cite{kipf2016semi}, GraphSAGE~\cite{hamilton2017inductive}, GAT~\cite{DBLP:conf/iclr/VelickovicCCRLB18}, and GraphSAINT~\cite{DBLP:conf/iclr/ZengZSKP20}. 
The idea behind EPT-based GNNs is similar to the conventional convolution: it passes the input signals through a set of filters to propagate the information, which is further followed by nonlinear transformations.
As a result, EP and ET operations are inherently intertwined in this pattern, i.e., each EP operation requires a neural layer to transform the hidden representations to generate the new embeddings for the next step. 
Motivated by ResNet~\cite{he2016deep}, some GNNs with the \textbf{EPT-SC} pattern deepen the EPT-based GNNs with skip connections, and both JK-Net ~\cite{xu2018representation}, and ResGCN ~\cite{li2019deepgcns} are the representative methods of this category. 

Besides the strict restriction that $D_p = D_t$, EPT-based and EPT-SC-based GNNs also suffer from low scalability and low efficiency. 
On the one hand, a deeper structure has more parameters, resulting in greater computation costs.
On the other hand, the number of nodes within each node's neighborhood grows exponentially with the increase of model depth in typical graphs, incurring significant memory requirement in a single machine or high communication costs in distributed environments~\cite{zheng2020distdgl}.

\para{Disentangled Graph Convolution.}
Previous works~\cite{frasca2020sign,DBLP:conf/sigir/0001DWLZ020,wu2019simplifying,zhang2021rod} have shown that the true effectiveness of GNNs lies in the EP operation rather than the ET operation inside the graph convolution.
Therefore, some disentangled GNNs propose to separate the ET operation from the EPT scheme. 
They can be classified into the following two categories according to the order of EP and ET operations.

One pattern is the Disentangled Transformation and Propagation (\textbf{DTP}). DTP-based GNNs firstly execute the ET operations and then turn to the EP operations, and the most representative methods in this category are PPNP~\cite{klicpera2018predict}, APPNP~\cite{klicpera2018predict}, and AP-GCN~\cite{spinelli2020adaptive}.
Unlike the DTP pattern, GNNs with the convolution pattern of Disentangled Propagation and Transformation (\textbf{DPT}) execute the EP operations in advance, and the propagated features are fed into a simple model composed of multiple ET operations~\cite{zhang2021gmlp}.
For example, SGC~\cite{wu2019simplifying} removes the nonlinearities in GCN and collapses all the weight matrices between consecutive layers into a simple logistic regression. 
Based on SGC, SIGN~\cite{frasca2020sign} further combines many graph convolutional filters in the EP operations, which differ in types and depths, utilizing significantly more structural information.

Compared with entangled GNNs, DTP-based GNNs can support large propagation depth $D_p$. Besides, DTP-based GNNs do not restrict $D_p=D_t$, enabling more flexibility in the GNN architecture design. 
However, they still have low scalability and efficiency issues when adapted to large graphs due to the high-cost propagation process during training.
On the contrary, DPT-based GNNs only need to precompute the propagated features once. Therefore, they are easier to scale to large graphs with less computation cost and lower memory requirement.

\section{Experimental Setup}
\label{sec.analysis}

In this section, we firstly introduce the evaluated datasets and models and then propose our smoothness measurement for analyzing the \textit{over-smoothing} issue. 

\para{Datasets and Models.}
To comprehensively evaluate previous GNNs, we select representative models according to their graph convolution patterns.
Concretely, we select GCN for EPT pattern, ResGCN and DenseGCN for EPT-SC pattern, DAGNN~\cite{liu2020towards} for DTP pattern, SGC, S$^2$GC~\cite{zhu2021simple} and Grand~\cite{feng2020graph} for DPT pattern, respectively.
The default dataset we use is PubMed~\cite{DBLP:journals/aim/SenNBGGE08}.
Besides, the Cora dataset~\cite{DBLP:journals/aim/SenNBGGE08} is used in Sec.~4.3, Sec.~6.1, and Sec.~6.2.
We also use a large dataset ogbn-arxiv~\cite{hu2020ogb} in Sec.~6.1.

\begin{figure*}[tp!]
\centering  
\subfigure[The influence of $D_p$ to model performance.]{
\label{fig.smooth_acc}
\includegraphics[width=0.30\textwidth]{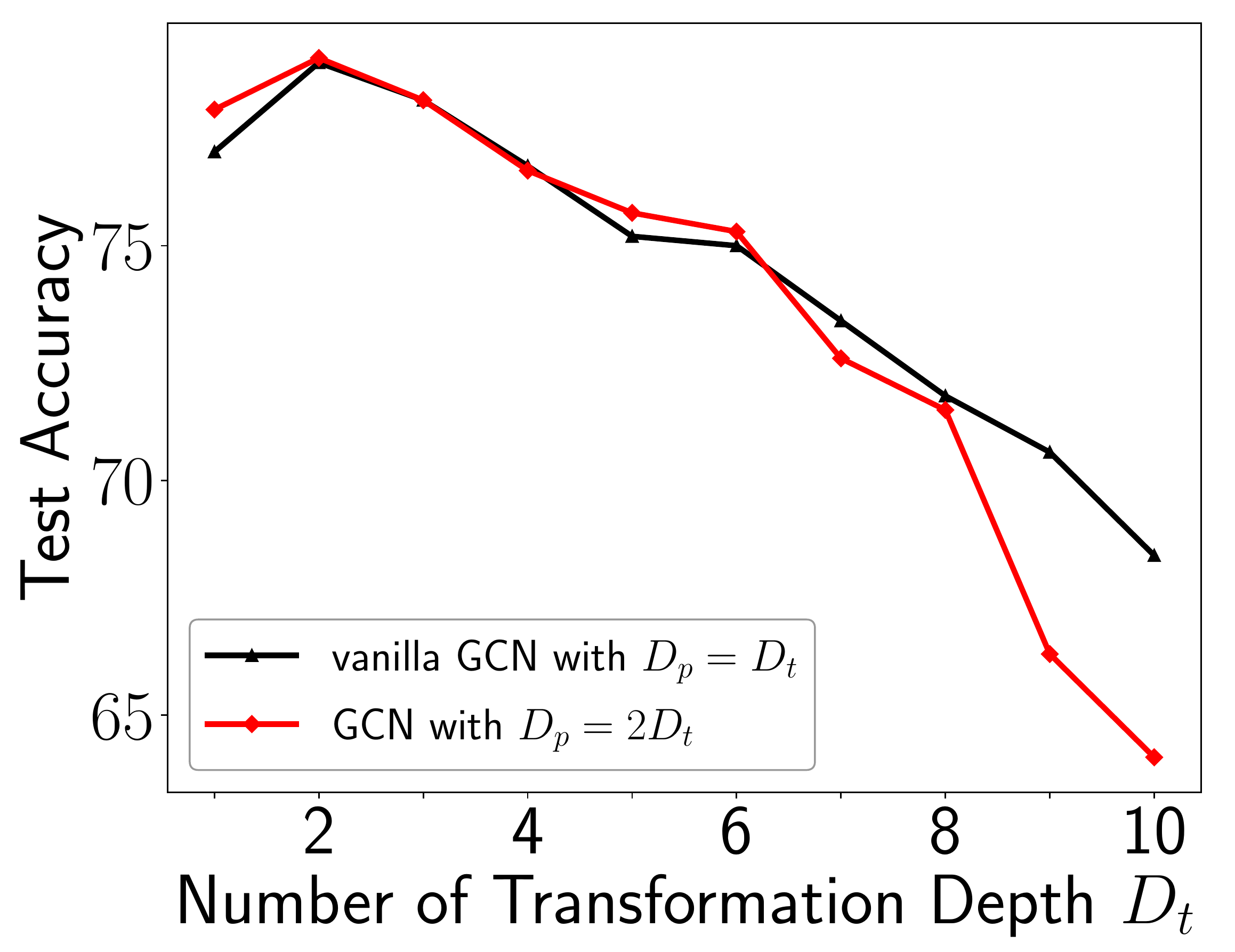}}
\subfigure[The influence of $GSL$ to model performance.]{
\label{fig.smooth_nsl}
\includegraphics[width=0.30\textwidth]{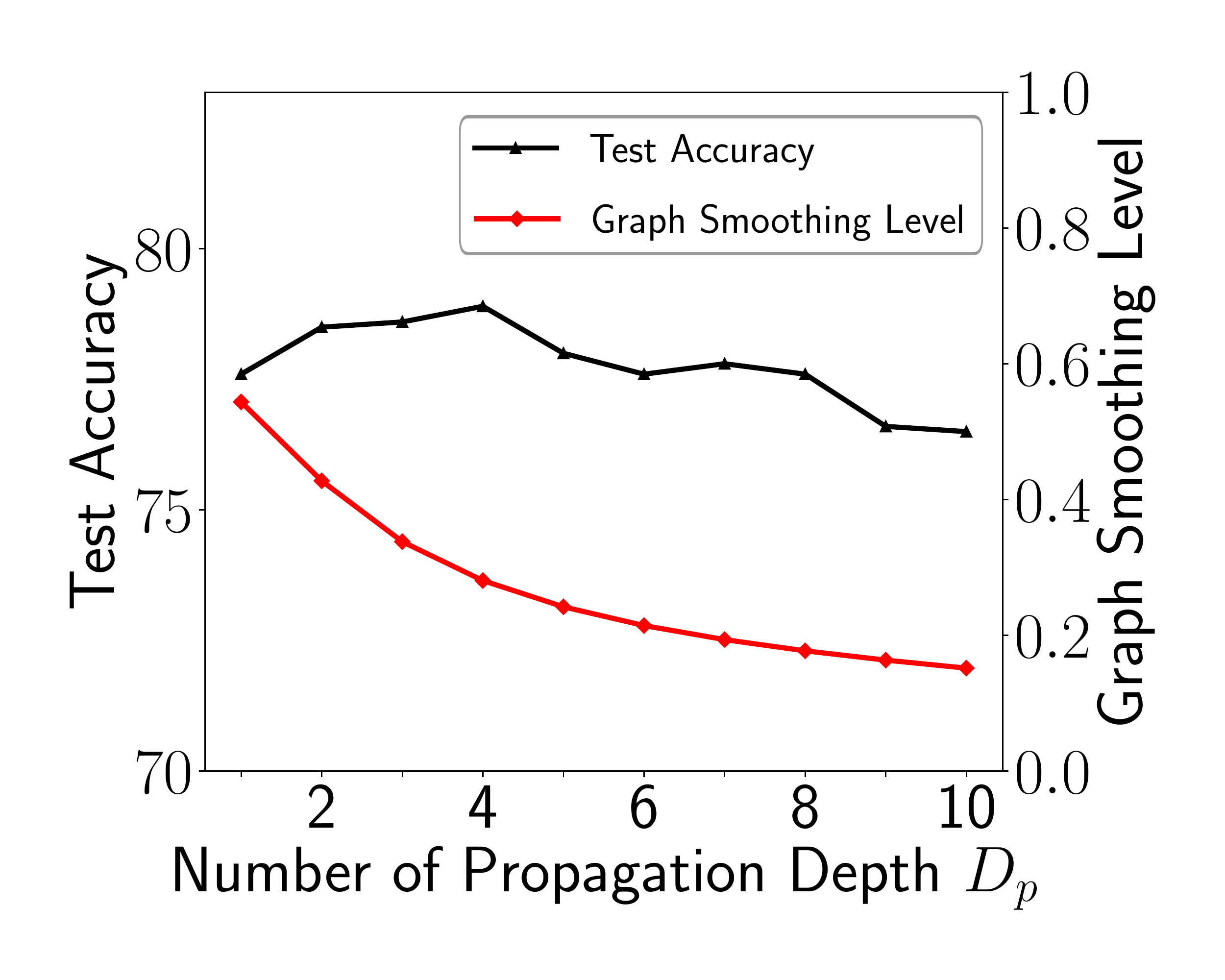}}
\subfigure[The influence of $D_t$ to model performance.]{
\label{fig.smooth_ak}
\includegraphics[width=0.30\textwidth]{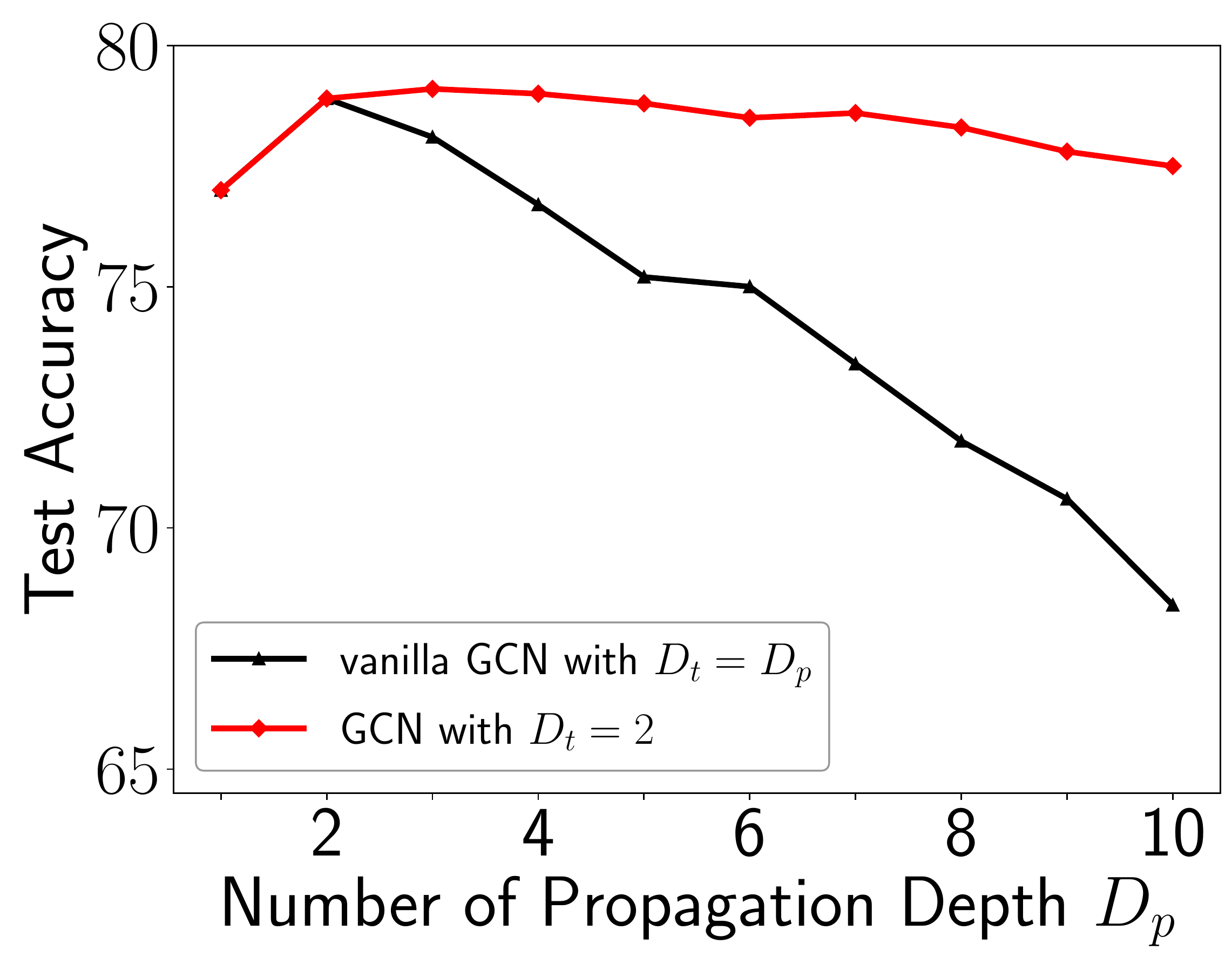}}
\caption{Over-smoothing is not the main contributor who hurts the performance of deep GNNs.}
\label{fig.oversmooth}
\end{figure*}

\para{Smoothness Measurement.}
\label{sec.smooth_metric}
As is shown in E.q.~\ref{eq_EP}, the normalized adjacency matrix $\mathbf{\hat{A}}$ is used in GNN to propagate each node's feature to its neighbors.
Each time $\mathbf{\hat{A}}$ multiplies with $\mathbf{X}$, the information one more hop away can be obtained.
Thus, in order to fully leverage high-order neighborhood information, a series of multiplications of $\mathbf{\hat{A}}\mathbf{X}$ have to be carried out, meaning stacking multiple EP operations.
However, if we execute $\mathbf{\hat{A}}\mathbf{X}$ for infinite times, the node representations within the same connected component would reach a stationary state.
Concretely, when applying $\widetilde{\mathbf{D}}^{r-1}\tilde{\mathbf{A}}\widetilde{\mathbf{D}}^{-r}$ as $\mathbf{\hat{A}}$, the stationary state follows
\begin{equation}
\small
\label{eq.stationary}
\hat{\mathbf{A}}^{\infty}_{i,j}  =  \frac{(d_i+1)^r(d_j+1)^{1-r}}{2M+N},
\end{equation}
where $N$ and $M$ are the number of nodes and edges in the graph, respectively.
It shows that after infinite steps propagation, the influence from node $i$ to node $j$ is only determined by the their degrees.
Under this scenario, the neighborhood information is thoroughly corrupted, resulting in awful node classification performance.
As the \textit{over-smoothing} issue is only introduced by the EP operation rather than the ET operation, here we introduce a new measurement ``Node Smoothness Level ($NSL$)'', to evaluate the smoothness of a node after $k$ steps propagation. Suppose $\mathbf{X}^{(0)} = \mathbf{X}$ is the original node feature matrix, and $\mathbf{X}^{(k)} = \mathbf{\hat{A}}^{k}\mathbf{X}^{(0)}$ is the smoothed features after $k$ steps propagation.

\begin{definition}[\textbf{Node Smoothing Level}]
\small
\label{df.nsl}
The Node Smoothing Level $NSL_{v}(k)$ parameterized by node $v$ and $k$ steps propagation is defined as: 
\begin{equation}
\label{eq.alpha}
\begin{aligned}
&\alpha = Sim(\mathbf{x}_v^{k},\mathbf{x}_v^{0}),\\
&\beta = Sim(\mathbf{x}_v^{k},\mathbf{x}_v^{\infty}),\\
NS&L_{v}(k) = \alpha * (1-\beta),
\end{aligned}
\end{equation}
where 
$\mathbf{x}_v^{k} \in \mathbf{X}^{(k)}$ is the smoothed feature of node $v$, and $\mathbf{x}_v^{\infty} \in \mathbf{X}^{(\infty)}$ represents node $v$'s feature at stationary state. 
$Sim(\cdot)$ is a similarity function, being the cosine similarity in the following discussion.
\end{definition}
Intuitively, $\alpha$ represents the similarity between a node's original feature and its representation after $k$ steps propagation.
On the other hand, $\beta$ represents the similarity between a node's stationary state and its current state after $k$ steps propagation.
Thus, when adopting $NSL_{v}(k) = \alpha * (1-\beta)$, we measure both the distance from node $v$ is to its original state and its stationary state after $k$ steps propagation.
Larger $NSL_{v}$ means that node $v$ is closer to its original state and farther away from its stationary state, in other words, less smoothed.

Further, the Graph Smoothing Level ($GSL$) is defined as:
\begin{equation}
\small
GSL(k) = \frac{1}{N}\sum_{v \in \mathcal{V}}NSL_{v}(k).
\end{equation}
Smaller $GSL(k)$ means that the graph $\mathcal{G}$ is more likely to forget the original node feature information $\mathbf{X}^{(0)}$ after $k$ steps propagation and has higher risks of the \textit{over-smoothing} issue.

\section{Misconceptions}
Although lines of works have managed to design deep GNN architectures, \textbf{what actually limits the stacking of EP and ET operations on graph-structure data} has not been well understood. Below, we empirically and experimentally analyze some misleading reasons and then propose our explanation.

\subsection{Over-smoothing}

Previous works~\cite{li2018deeper, zhang2019attributed} show that the EP operation in GNNs is exactly a process of Laplacian smoothing, which smooths adjacent node embeddings. That is to say, embeddings of some nodes will converge to the same value when GNN goes deeper. 
Therefore, most current researches owe the performance degradation of deep GNNs to \textit{over-smoothing}.
For example, Grand~\cite{feng2020graph} randomly drops raw features of nodes before propagation, making node embeddings more distinguishable, while DropEdge~\cite{rong2019dropedge} randomly removes edges during training. 
Despite their ability to go deeper with better node classification performance, the explanations for their effectiveness are misleading in some instances.

To investigate the relations between graph smoothness and the node classification accuracy, we increase the number of graph convolutional layers in vanilla GCN ($D_p = D_t$) and a modified GCN with $\hat{\mathbf{A}}^{2}$ being the normalized adjacency matrix ($D_p = 2D_t$) on the PubMed dataset.
The results are shown in Fig.~\ref{fig.smooth_acc}.

\para{Finding 1.} \textbf{Even with a higher level of smoothness, GCN with $D_p = 2D_t$ always has similar test accuracy with vanilla GCN ($D_p = D_t$) when $D_t$ ranges from 1 to 8, and the \textit{over-smoothing} seems to begin dominating the performance decline only when $D_p$ exceeds 16 ($2 \times 8$).}
The performance of vanilla GCN does decrease sharply when $D_p$ exceeds 2, which is precisely the situation \textit{over-smoothing} suggests.
However, the test accuracy of the model with larger smoothness ($D_p = 2D_t$) is similar to the vanilla GCN, which on the contrary implies that \textit{over-smoothing} may not be the major cause for performance degradation of deep GCNs even when $D_p$ is relatively large (e.g., 16).

\para{Implication.}\textit{Over-smoothing} may not be the major factor that limits the model depth of GNN until it achieves an extremely high level (i.e., $D_p \geq 16$ in the PubMed dataset).

To further validate this guess, we increase the number of propagation depth $D_p$ of SGC and then report the corresponding test accuracy and the $GSL$, the smoothness measurement proposed in Sec.3.1, in Fig.~\ref{fig.smooth_nsl}.

\para{Finding 2.} \textbf{By increasing $D_p$ from 1 to 10, the $GSL$ has decreased by more than 60\%, but the corresponding test accuracy decline is less than 1\%}. 
The model seems to be affected by \textit{over-smoothing} when $D_p$ exceeds 4 due to the low graph smoothing level. However, compared with 10-layer GCN in Fig.~\ref{fig.smooth_acc}, the corresponding test accuracy of 10-layer SGC is still high even it has the same $D_p$ with GCN.
Therefore, we further guess that the large $D_t$ in vanilla GCN may be the main cause for the performance degradation of deep GCNs.

\para{Implication.}The influence of \textit{over-smoothing} is overstated, and the performance degradation of deep GNNs may be dominated by \textit{model degradation}.

To dig out the true limitation of deep GCNs, we fix the number of transformation depth $D_t$ to $2$ and set the normalized adjacency matrix to $\hat{\mathbf{A}}^{D_p/2}$ (when $D_p$ is odd, use $\hat{\mathbf{A}}^{\lfloor D_p/2 \rfloor + 1}$ in the first layer and $\hat{\mathbf{A}}^{\lfloor D_p/2 \rfloor}$ in the second layer), and then report the test accuracy along with the increased propagation depth $D_p$.

\para{Finding 3.} \textbf{The test accuracy of GCN with $D_t = 2$ does not drop quickly when $D_p$ becomes large, while it faces a sharp decline in vanilla GCN, which fixes $D_p = D_t$.} Individually enlarging $D_p$ will increase the level of \textit{over-smoothing}, but the test accuracy is only slightly influenced. However, the performance of GCN experiences a drastic drop if we simultaneously increase $D_t$.

\para{Implication.}The \textit{model degradation} issue introduced by large $D_t$ is the major limitation of deep GNNs in this scenario.

\begin{figure}
	\centering
	\includegraphics[width=.63\linewidth]{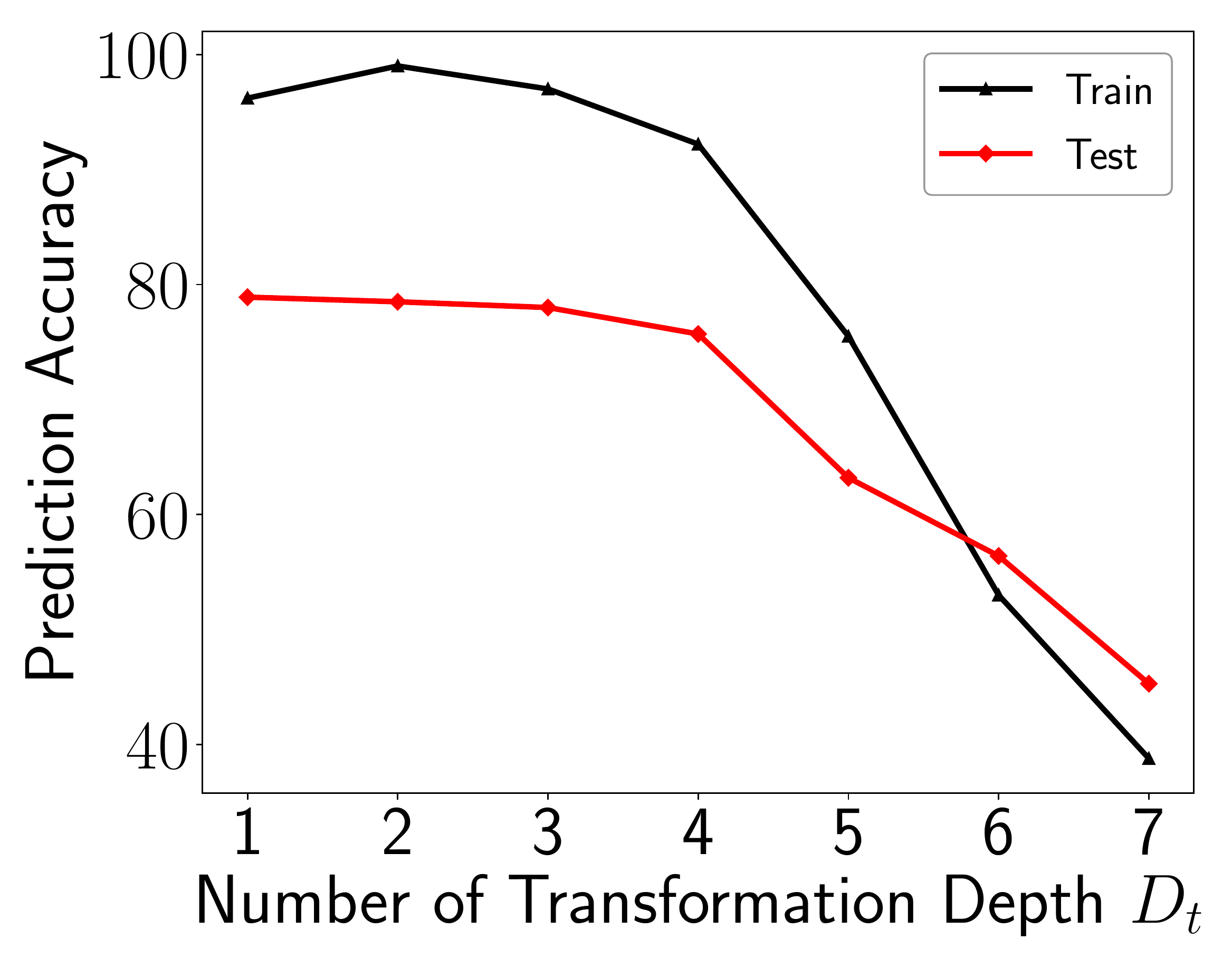}
	\caption{Training and test accuracy both drop sharply when model grows deep.}
	\label{fig.overfit}
\end{figure}

\subsection{Over-fitting}
Some works~\cite{rong2019dropedge,li2019deepgcns,zhou2020understanding,yang2020revisiting} attribute the performance degradation of deep GNNs to over-fitting. Concretely, over-fitting comes from the case when an over-parametric model tries to fit a distribution with limited training data, which results in learning well on the training data but failing to generalize to the testing data. 
We plot the corresponding node classification accuracy of GCN on both the training and the test set under different model depths in Fig.~\ref{fig.overfit}.

\para{Finding.} \textbf{Both the training and test accuracy drop quickly when the model grows deep.} However, over-fitting assumes that over-parametric models get low training error but high testing error, inconsistent with the experimental results.

\para{Implication.}Over-fitting is not the primary cause for performance degradation of deep GNNs.

\begin{figure}
	\centering
	\includegraphics[width=.63\linewidth]{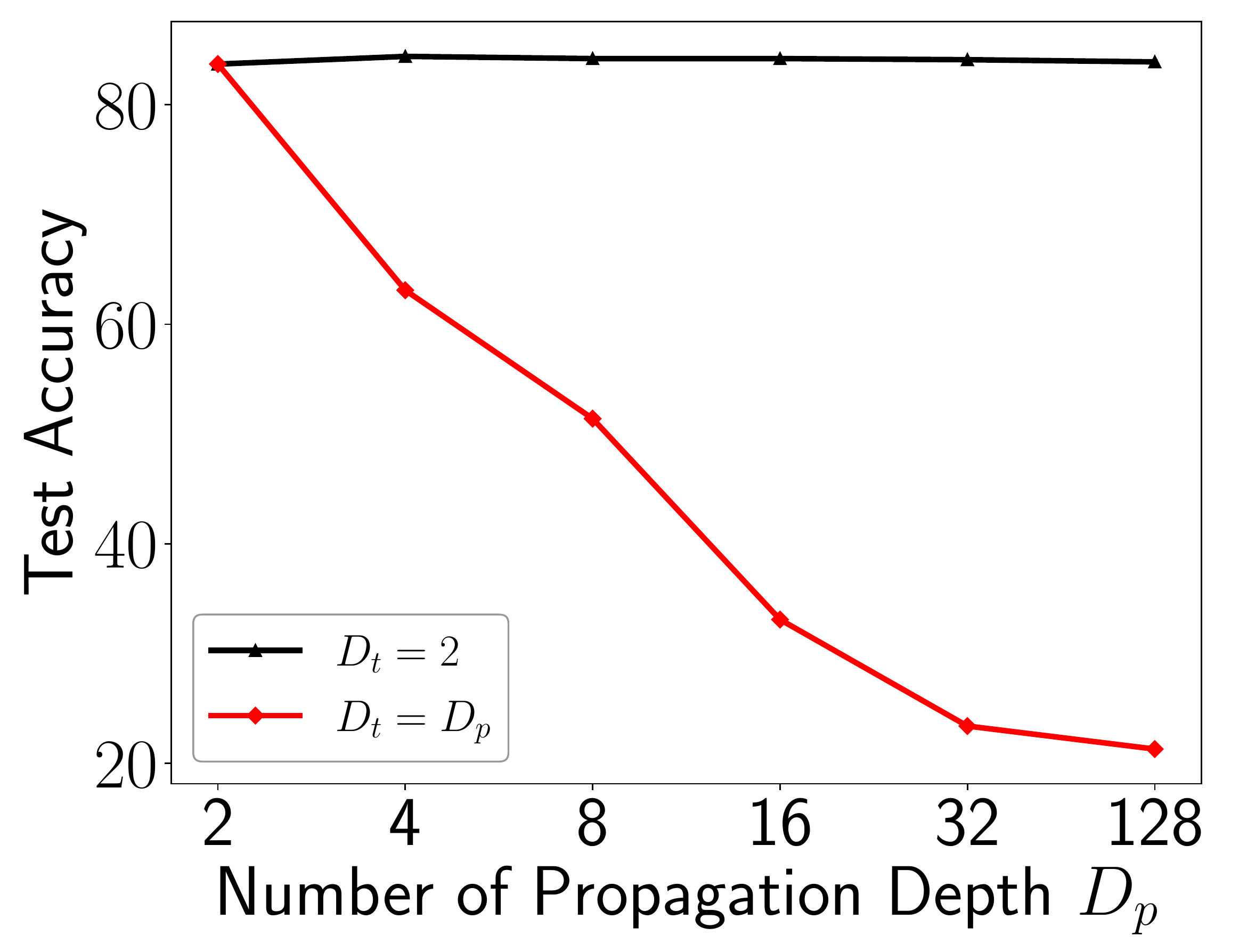}
	\caption{Fixing $D_t=D_p$ degrades the performance badly when $D_p$ becomes large on Cora dataset.}
	\label{fig:coupled}
\end{figure}

\subsection{Entanglement} 
Some recent works~\cite{DBLP:conf/sigir/0001DWLZ020, liu2020towards} argue that the key factor compromising the performance of deep GNNs is the entanglement of EP and ET operations in current graph convolutional layers. 
For example, DAGNN claims that the entanglement of EP and ET operations in GCN is the true reason it only supports shallow architectures. 
To investigate this statement, we vary the transformation depth $D_t$ in ResGCN, DenseGCN, and vanilla GCN and report their test accuracy on the PubMed dataset.
The experimental results are shown in Fig.~\ref{fig.res_acc}.

\para{Finding 1.} \textbf{Although ResGCN and DenseGCN have entangled designs, when $D_p$ and $D_t$ both become large, they do not experience significant performance drop as GCN does.} Both the residual and dense connections can effectively alleviate the influence of \textit{model degradation}, and the performance degradation of ResGCN and DenseGCN starting at $D_t=6$ might be caused by the \textit{over-smoothing} issue.

\para{Implication.}GNNs can go deep even in an entangled design, and the entanglement of EP and ET operations may not be the true limitation of the GNN depth.

What is worth noting is that previous works~\cite{zhu2021simple, liu2020towards}, which have disentangled designs and state that they support deep architectures, are only able to go deep on the propagation depth $D_p$.
In their original design, if we increase $D_t$, their performance will also degrade badly.
To validate this, we run DAGNN in two different settings: the first controls $D_t=2$ and increases $D_p$, the second controls $D_t=D_p$ and increases $D_p$.
The test accuracy under these two settings on the PubMed dataset is shown in Fig.~\ref{fig:coupled}.

\para{Finding 2.} \textbf{The disentangled DAGNN also experiences huge performance degradation when the model owns large $D_t$.} If the entanglement dominates the performance degradation of deep GNNs, DAGNN should be able to also go deep on $D_t$. However, if we individually increase $D_t$ of DAGNN, the sharp performance decline still exists.

\para{Implication.}The major limitation of deep GNNs is the \textit{model degradation} issue introduced by large $D_t$ rather than the entanglement of EP and ET operation.

\begin{figure}
	\centering
	\includegraphics[width=.63\linewidth]{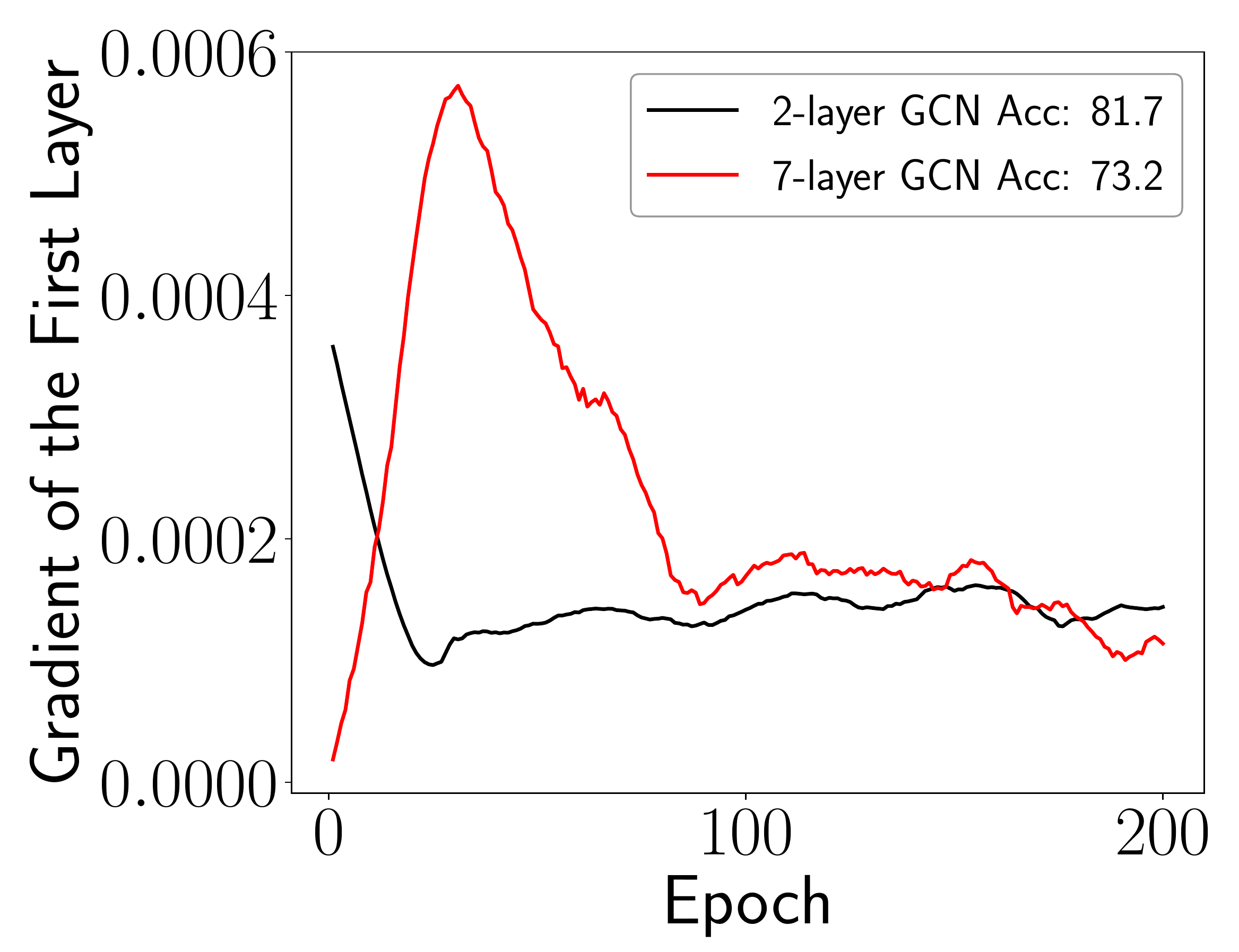}
	\caption{First layer gradient comparison of GCN with different layers on Cora dataset.}
	\label{fig:gradient}
\end{figure}

\subsection{Gradient Vanishing}
Gradient vanishing means that the low gradient in the shallow layers makes it hard to train the model weights when the network goes deeper, and it has a domino effect on all of the further weights throughout the network. 
To evaluate whether the gradient vanishing exists in deep GNNs, we respectively perform node classification experiments on the Cora dataset and plot the gradient -- the mean absolute value of the gradient matrix of the first layer in the 2-layer and 7-layer GCN in Fig.~\ref{fig:gradient}.

\para{Finding 1.} \textbf{Although the test accuracy of 7-layer GCN drops quickly, its gradient is as large as the gradient of 2-layer GCNs, or even larger in the initial model training phases.}
The explanation for the initial gradient rise of the 7-layer GCN might be that the large model needs more momentum first to adjust and then jump out of the suboptimal local minima initially.

\para{Implication.}Gradient vanishing is not the leading cause of performance degradation in deep GNNs.

\section{Root Causes}
In this section, we first analyze why some entangled and disentangled GNNs can go deeper while maintaining similar or better model performance as shallow architectures. Then, we summarize some common insights.

\subsection{Entangled Convolution}
The above findings show that previous works tackling the shallow architecture issue in GNNs are misleading.
It motivates us to dig deeper into the true reasons of the performance degradation of some deep GNNs. 
As we discussed in Sec.~2.2, GCN is an augmented MLP where each node uses the features of neighboring nodes to enhance its representation. However, if the base MLP fails to extract information, individually improving the upstream propagation operation cannot boost the performance either. Therefore, \textbf{if deep MLP fails, how can we expect that deep GNNs can work?}

\begin{figure}[tp!]
\centering  
\subfigure[The skip connection to MLP.]{
\label{fig.mlp}
\includegraphics[width=0.22\textwidth]{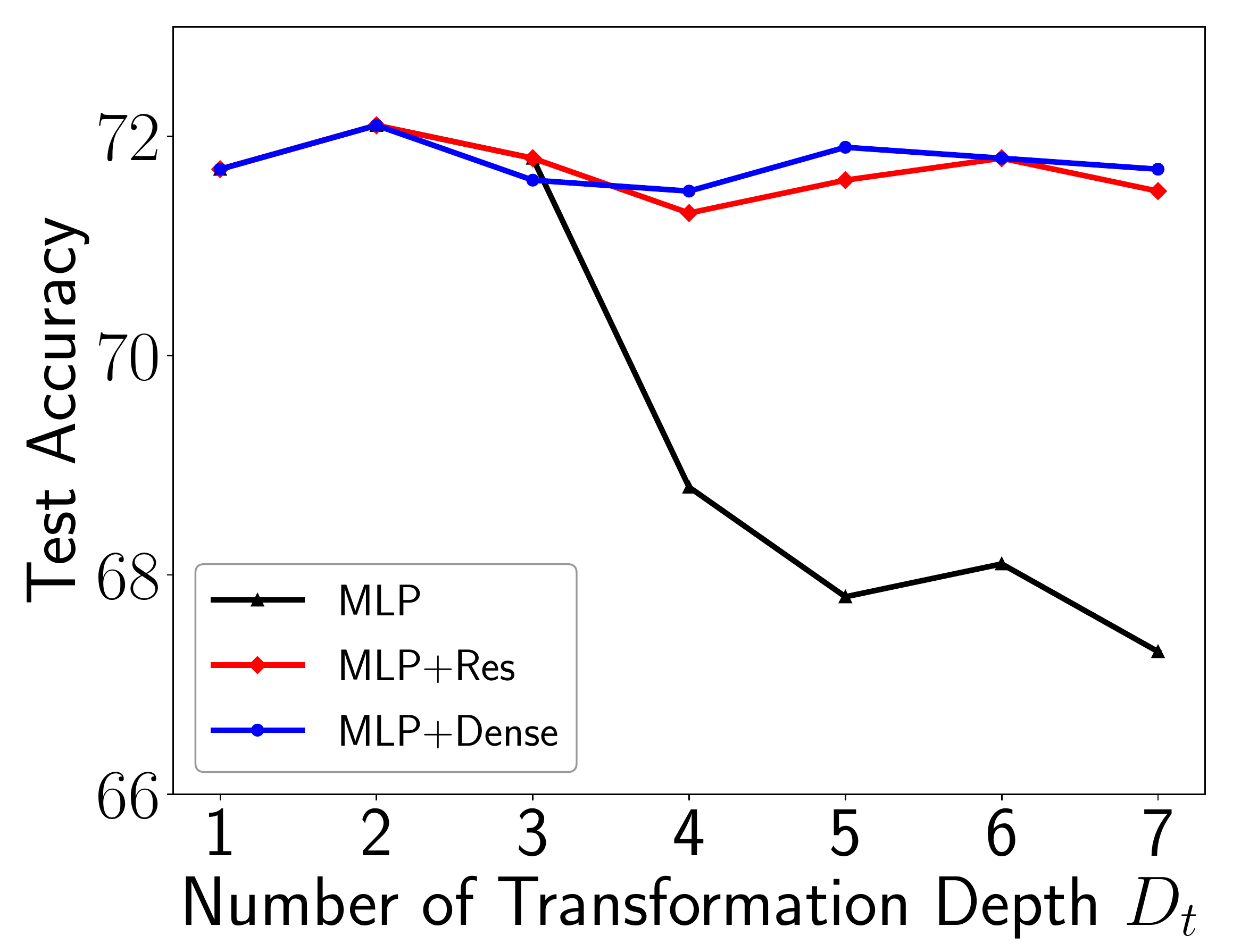}}\hspace{1mm}
\subfigure[The skip connection to GCN.]{
\label{fig.res_acc}
\includegraphics[width=0.22\textwidth]{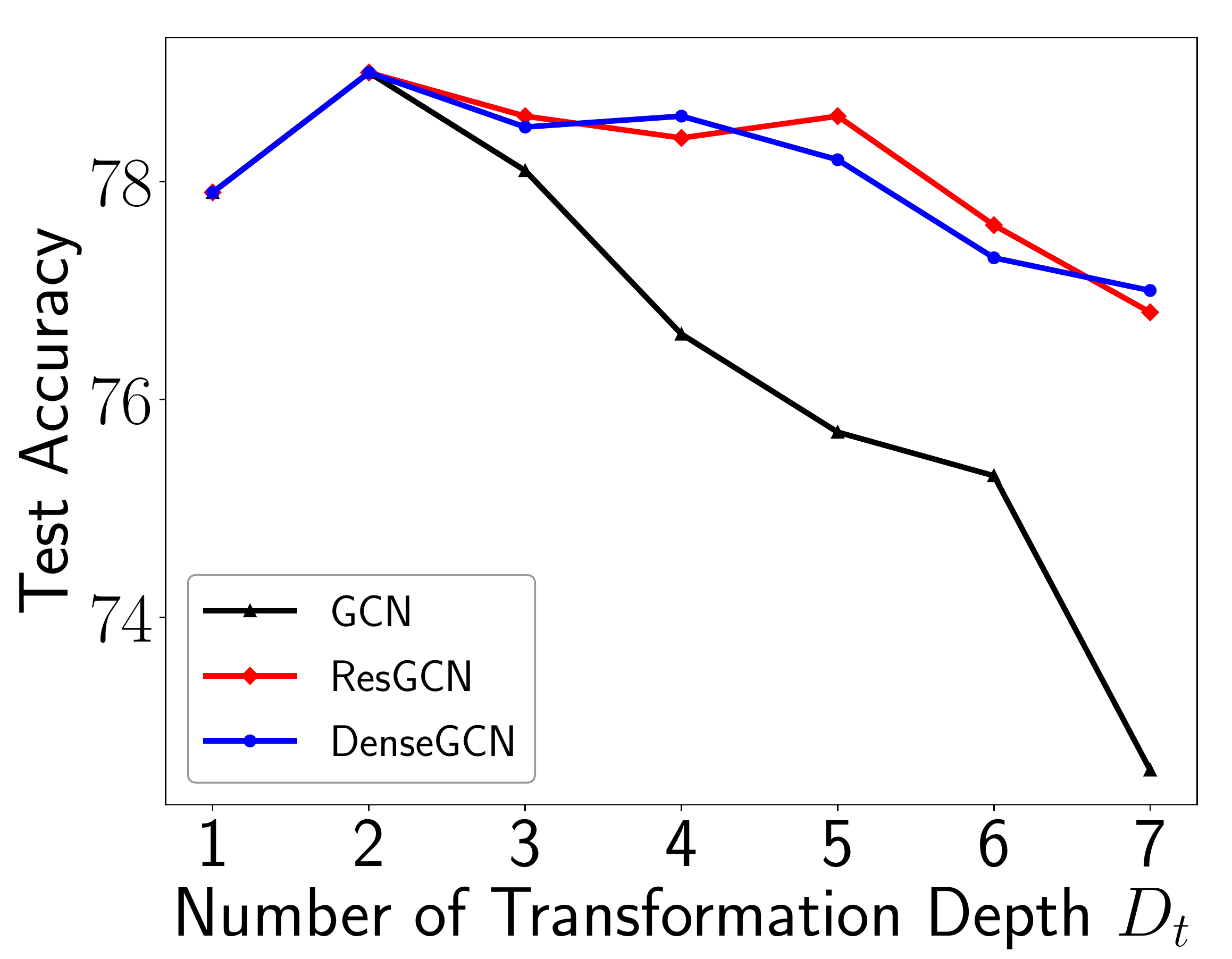}}\hspace{1mm}
\caption{Performance comparison when adding Residual and Dense connection.}
\label{fig.res_dense}
\end{figure}

To measure the influence of ET operation in deep GNNs, We remove the EP operations in GCN, degrading it to a plain MLP. Moreover, we also add the residual or dense connections to it and generate two MLP variants: ``MLP+Res'' and ``MLP+Dense''. The test accuracy of these models with increasing $D_t$ is shown in Fig.~\ref{fig.mlp}.

\para{Finding 1.} \textbf{The test accuracy of MLP drops quickly from $D_t$ = 3 to $D_t$ = 7, while the performance becomes stable when residual/dense connections are added.} 
Equipped with residual/dense connections, the \textit{model degradation} issue of deep MLPs can be effectively alleviated.

\para{Implication.}\textit{Model degradation} introduced by large $D_t$ is the main cause for the performance degradation of deep entangled GNNs.

To validate this, we further increase the model depth of ResGCN and DenseGCN~\cite{li2019deepgcns} ($D_p = D_t$) and test their performance on the PubMed dataset. 

\para{Finding 2.} \textbf{Fig.~\ref{fig.res_acc} shows that the residual/dense connection can alleviate the performance degradation of deep GCNs.}  Although the \textit{over-smoothing} issue occurs and the performance of ResGCN and DenseGCN decreases when $D_p$ is larger than 5, the decrease is much slighter than the one of GCN.

\para{Implication.}\textit{Model degradation} dominates the performance degradation before \textit{over-smoothing} in deep entangled GNNs.

This surprising finding differs from the prevailing belief that over-smoothing, over-fitting, the entanglement of EP and ET, or gradient vanishing is the leading cause for the performance degradation of deep GNNs.
To sum up, both the \textit{over-smoothing} introduced by large $D_p$ and the \textit{model degradation} introduced by large $D_t$ are the causes for the compromised performance of deep GNNs, and \textit{model degradation} may have stronger influence.

\subsection{Disentangled Convolution}
\begin{figure}[tp!]
\centering  
\subfigure[Test accuracy when $D_p$ increases.]{
\label{fig.deepnofail_dp}
\includegraphics[width=0.22\textwidth]{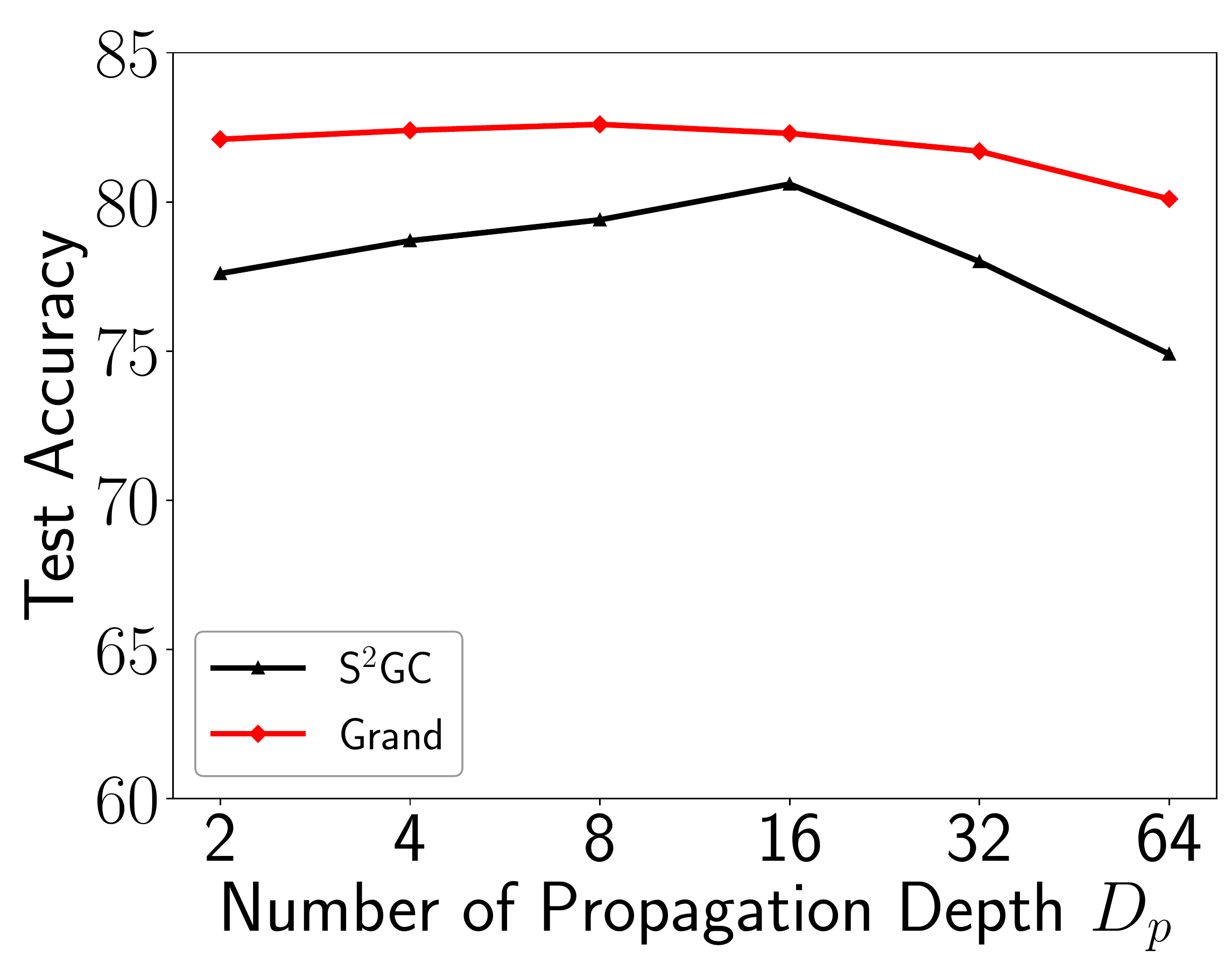}}\hspace{1mm}
\subfigure[Test accuracy when $D_t$ increases.]{
\label{fig.deepnofail_dt}
\includegraphics[width=0.22\textwidth]{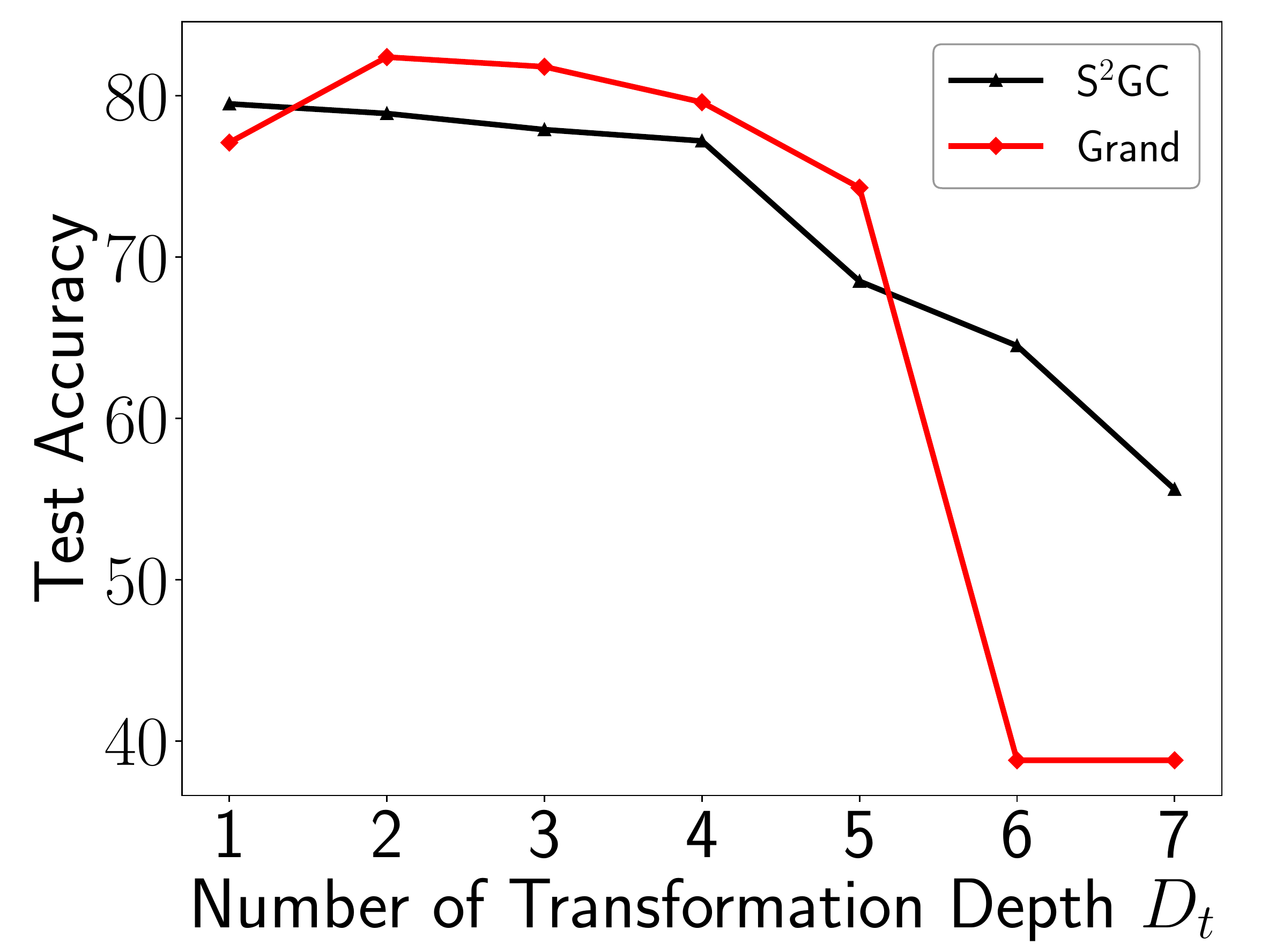}}\hspace{1mm}
\caption{Deep disentangled GNNs cannot go deep on $D_t$.}
\label{fig.deepnofail}
\end{figure}

We then investigate \textbf{why disentangling EP and ET is able to allow more EP operations. }
Concretely, we carefully investigate current disentangled GNNs~\cite{frasca2020sign,DBLP:journals/corr/abs-2004-11198} and find that the decoupling strategy makes the propagation and the transformation operations independent, thus $D_p$ and $D_t$ are not forced to be the same. Therefore, disentangled GNNs generally fix $D_t$ and increase $D_p$ to capture deeper graph structural information.
Here we select two disentangled GNNs, S$^2$GC and Grand, which state that they support deep architectures. Their performance of individually increasing $D_p$ and $D_t$ are shown in Fig.~\ref{fig.deepnofail_dp} and Fig.~\ref{fig.deepnofail_dt}, respectively.

\para{Finding.} \textbf{Individually increasing $D_p$ would not incur a severe performance drop even when $D_p$ is increased to 64, while their performance would experience a sharp decline when $D_t$ increases.} Increasing $D_p$ enlarges each node's receptive field, thus leading to more decent representations, which is demonstrated to be even more beneficial for sparse graphs in our following experiments (See Sec.~6.1).

\para{Implication.} Deep disentangled GNNs are flexible to individually increase $D_p$, so that the \textit{model degradation} introduced by large $D_t$ can be avoided. 

\subsection{Summary}
Based on the analysis of deep entangled and disentangled GNNs, we provide the following observations: 
\begin{itemize}
\item GNNs have two model depths: the number of EP operations $D_p$ and the number of ET operations $D_t$.
\item Larger $D_p$ and $D_t$ will increase the level of \textit{over-smoothing} and \textit{model degradation}, respectively. Besides, \textit{model degradation} dominates the performance degradation before \textit{over-smoothing} on most datasets as GNNs go deeper; 
\item The skip connections in EPT-SC-based GNNs can alleviate the \textit{model degradation} issue; 
\item Most disentangled GNNs only increase $D_p$, thus avoid the occurrence of the \textit{model degradation} issue. 
\end{itemize}

\begin{table}[]
\centering
{
\noindent
\caption{Test accuracy under different edge sparsity. }
\label{table.edge_sp}
\resizebox{0.95\linewidth}{!}{
\begin{tabular}{ccccccccc}
\toprule
\textbf{Datasets} & \textbf{$\frac{M}{N^2}$} & \textbf{2} & \textbf{4} & \textbf{6} & \textbf{8} & \textbf{12} & \textbf{16} & \textbf{20} \\ \midrule
Cora & 0.7\textperthousand & \textbf{59.8}  & 59.6 & 57.9 & 57.3 & 56.5 & 51.8 & 47.1   \\ \midrule
PubMed  & 0.1\textperthousand & 78.5  & \textbf{78.9}  & 79.4  & 77.6  & 77.3 & 76.6  & 75.8  \\ \bottomrule
\end{tabular}}}
\end{table}

\begin{table}[tpb!]
\centering
{
\noindent
\caption{Test accuracy under different label missing rates.}
\label{table.label_sp}
\renewcommand{\multirowsetup}{\centering}
\resizebox{0.95\linewidth}{!}{
\begin{tabular}{ccccccccc}
\toprule
\textbf{Datasets}&\textbf{Labels/class}&\textbf{2}& \textbf{4}&\textbf{6}&
\textbf{8}&\textbf{12}&\textbf{16}&\textbf{20}\\
\midrule
\multirowcell{2}{Cora}&
20& \textbf{81.5}  & 81.3 & 80.8 & 80.1 & 80.0 & 79.5 & 78.8  \\
&1& 53.8  & 59.2 & 62.9 & 64.3 & 66.5 & \textbf{66.7} & 65.6   \\
\midrule
\multirowcell{2}{PubMed}&
20& 78.5  & \textbf{78.9}  & 79.4  & 77.6  & 77.3 & 76.6  & 75.8 \\
&1& 61.2  & 65.9  & 67.4  & 67.5  & 67.7 & \textbf{69.0}  & 68.3 \\
\bottomrule
\end{tabular}}}
\end{table}

\begin{table}[tpb!]
\centering
{
\noindent
\caption{Test accuracy under different feature missing rates.}
\label{table.feat_sp}
\renewcommand{\multirowsetup}{\centering}
\resizebox{0.95\linewidth}{!}{
\begin{tabular}{ccccccccc}
\toprule
\textbf{Datasets}&\textbf{Feature missing \%}&\textbf{2}& \textbf{4}&\textbf{6}&
\textbf{8}&\textbf{12}&\textbf{16}&\textbf{20}\\
\midrule
\multirowcell{2}{Cora}&
0\%& \textbf{81.5}  & 81.3 & 80.8 & 80.1 & 80.0 & 79.5 & 78.8  \\
&50\%& 75.4  & 77.7 & 78.5 & \textbf{79.5} & 79.4 & 78.9 & 78.0   \\
\midrule
\multirowcell{2}{PubMed}&
0\%& 78.5  & \textbf{78.9}  & 79.4  & 77.6  & 77.3 & 76.6  & 75.8 \\
&50\%& 60.6  & 65.1  & 66.3  & 66.7  & 68.7 & \textbf{69.2}  & 68.7 \\
\bottomrule
\end{tabular}}}
\end{table}

\section{Insights and Guidelines}
\subsection{When We Need Deep GNN Architectures?}
\para{When we need deep EP?}
We review the related works and researches concerning $D_p$'s importance and then investigate experimentally when it is appropriate to enlarge $D_p$ in GNNs. 
GNNs mainly benefit from performing EP operations over neighborhoods. Stacking more EP operations will expand a node's receptive field and help it to gain more deep structural information. 
However, if the graph is dense, increasing $D_p$ may lead to \textit{over-smoothing}, which results from the rapid expansion of the receptive field.

Generally, most real-world graphs exhibit sparsity in three aspects: edges, labels, and features. We define edge sparsity, label sparsity, and feature sparsity as follows:
 (1) \underline{Edge sparsity}: 
 nodes in real-world graphs usually have a skewed degree distribution, and many nodes are rarely connected~\cite{kuramochi2005finding}.
 (2) \underline{Label sparsity}: only a small part of nodes are labeled due to the high labeling costs or long labeling time~\cite{garcia2017few}.
 (3) \underline{Feature sparsity}: some nodes in the graph do not own features, i.e., the new users or products in a graph-based recommendation system~\cite{DBLP:conf/iclr/ZhaoA20}.
 We argue that \textbf{sparse graphs naturally need deeper EP for larger receptive fields}. 
 Experiments are conducted to demonstrate that the graph sparsity mentioned above will highly affect the optimal choice of $D_p$.
 SGC is adopted as the base model throughout these experiments.
 
Firstly, we investigate how edge sparsity influences the optimal $D_p$. For a fair comparison, we only sample part of the labels for Cora to have the same feature and label missing rates as PubMed. Then we increase $D_p$ from 2 to 20 on Cora and PubMed and report the test accuracy under each setting.
The experimental results in Fig.~\ref{table.edge_sp} show that the optimal $D_p$ is 4 for PubMed and 2 for Cora, meaning that the sparser graph requires a larger $D_p$, which further demonstrate the benefits of deepening $D_p$ under edge sparse conditions.
Secondly, we illustrate the relationship between the label sparsity and $D_p$. As shown in Fig.~\ref{table.label_sp}, we fix one labeled node per class on Cora and PubMed and keep the same setting with the edge sparsity experiment.
It can be seen that the classification accuracy increases as $D_p$ ascends from 2 to 16. However, if we increase the label rate to 20 nodes per class, the node classification drops from 2 layers. The experiment on PubMed also shows that graphs with lower label rates require larger $D_p$.   
Finally, in Table~\ref{table.feat_sp}, we report the test accuracy under the feature sparsity setting where 50\% node features are randomly dropped from Cora and PubMed datasets.
The result is consistent with the former two experiments that graphs with higher feature sparsity levels require larger $D_p$.


\begin{table}[]
\centering
{
\noindent
\caption{Replacing the logistic regression in SGC with ResMLP. The test accuracy of the corresponding model on small graph Cora and large graph ogbn-arxiv when the MLP depth $D_t$ changes from 1 to 7.}
\label{table.large_dt}
\resizebox{0.95\linewidth}{!}{
\begin{tabular}{cccccccc}
\toprule
\textbf{Dataset} & \textbf{1} & \textbf{2} & \textbf{3} & \textbf{4} & \textbf{5} & \textbf{6} & \textbf{7} \\ \midrule
Cora & 80.9 & \textbf{81.7}  & 81.5 & 81.1  & 80.7  & 80.5 & 80.2 \\ \midrule
ogbn-arxiv  & 70.1 & 70.2  & 71.2  & 71.4  & 71.4  & \textbf{71.6} & 71.3  \\ \bottomrule
\end{tabular}}}
\vspace{1mm}
\end{table}

\para{When we need deep ET?}
Generally, stacking multiple ET operations can better fit and learn the data distribution.  
However, we observe that the optimal $D_t$ for GNN is sensitive to the graph size. 
Concretely, \textbf{small graphs contain limited information, and shallow transformation architecture is enough for generating decent node embeddings. However, large graphs have complex structural information as complexity grows at a squared rate, which requires larger $D_t$ to extract meaningful information.}
We prove this insight by experimentally examining how $D_t$ influences the node classification performance on Cora and ogbn-arxiv datasets. 
Concretely, we replace the logistic regression in SGC with ``MLP+Res'' and increase $D_t$ from 1 to 7. 
Table~\ref{table.large_dt} shows that the accuracy on ogbn-arxiv increases as $D_t$ ascends from 1 to 6. However, $D_t = 2$ is enough for the small graph Cora.
Therefore, larger graphs need larger $D_t$ for high-quality node embeddings.

\begin{figure}
	\centering
	\includegraphics[width=.63\linewidth]{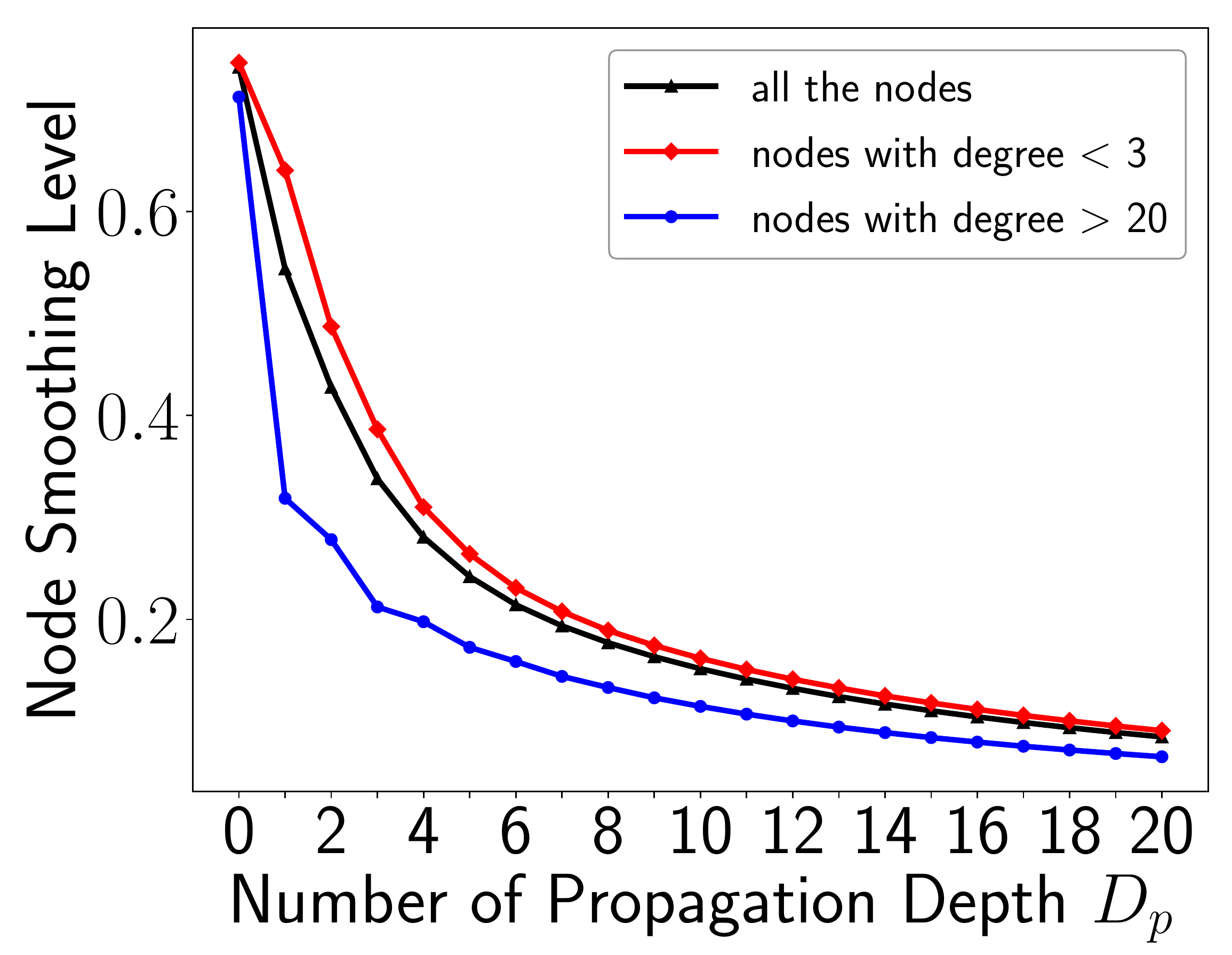}
	\caption{The average \textit{NSL} of each node group on PubMed.}
	\label{fig:smooth_speed}
\end{figure}

\subsection{How to Construct Deep GNN Architectures?}
\para{How to get deep EP?}It has been observed that sparser graphs require larger $D_p$.
Most current deep GNNs are designed for deeper embedding propagations, in other words, larger $D_p$. 
Concretely, APPNP~\cite{klicpera2018predict} proposes to utilize a propagation scheme derived from personalized PageRank~\cite{page1999pagerank} and propagates over multiple $k$-hop neighborhoods.
S$^2$GC~\cite{zhu2021simple} performs the propagation operations with generalized PageRank previously and gets weighted average representations of propagated features at different steps.  GBP~\cite{chen2020scalable} uses a modified Markov Diffusion Kernel for EP operation, which assigns the highest weight to the closest neighborhoods and smaller weights to nodes with more diffusion steps. 

\begin{figure}
	\centering
	\includegraphics[width=.9\linewidth]{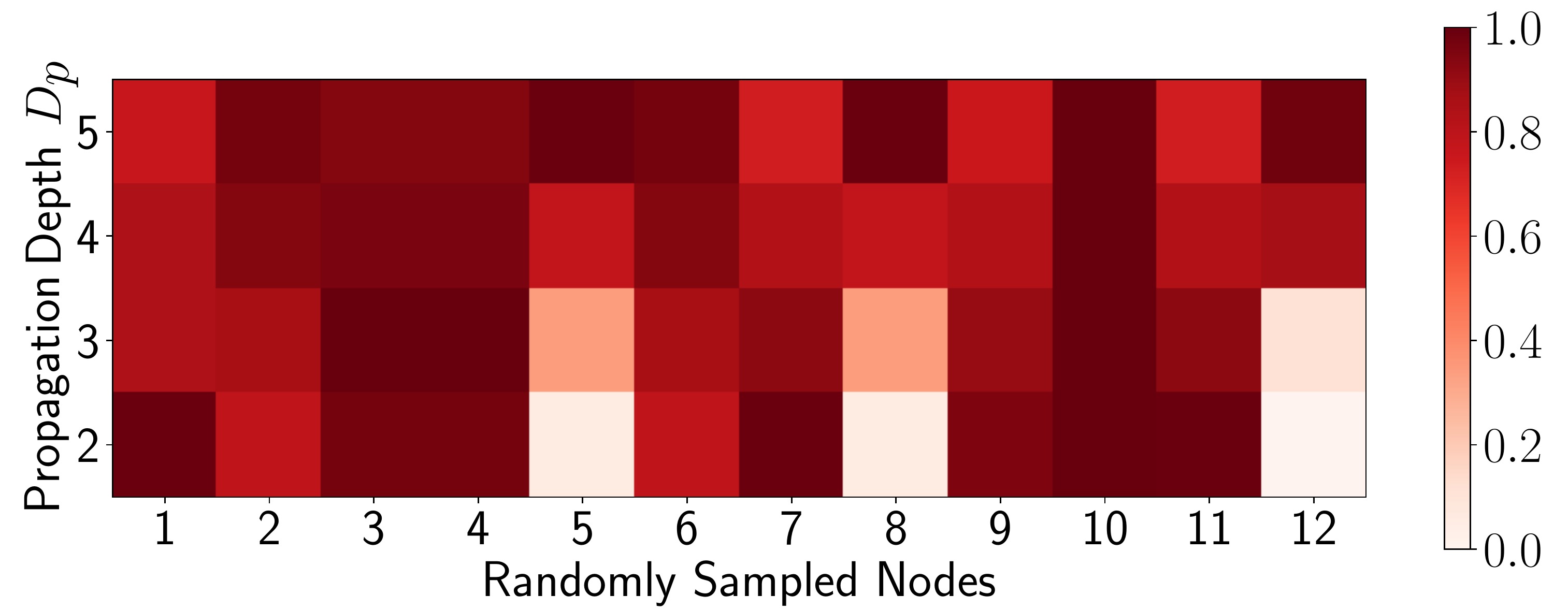}
	\caption{Different nodes reach their optimal performance at varied propagation step.}
	\label{fig:smooth_acc}
\end{figure}

Despite the effectiveness of previous works, there still exists a problem: the propagated features at different steps are combined at the graph level, which ignores the characteristics of certain nodes.
For example, the weighting scheme in GBP is inflexible because it assigns uniform weights to all the nodes when combining the propagated features at different propagation depths. 
To better explain this inflexibility, we divide the nodes on the PubMed dataset into three different groups according to their degrees and report the average $NSL$ of each node group.
The experimental results in Fig.~\ref{fig:smooth_speed} reveal that the nodes with larger degrees have larger smoothing speed and are more sensitive to the \textit{over-smoothing} issue than nodes with smaller degrees.
Therefore, we argue that \textbf{nodes at different positions of a graph require different $D_p$ to fully capture the structural information.}
To verify this, we apply SGC for node classification with different propagation depths on 12 randomly selected nodes of the Cora dataset. We run SGC 100 times and report the average accuracy of each node. We observe from Fig.~\ref{fig:smooth_acc} that the optimal propagation steps of these nodes are highly different. It demonstrates that $D_p$ should be flexible at the node level (i.e., different nodes are assigned with personalized weights). Our proposed DGMLP achieves this goal by propagating the node features to different steps and \textbf{combining them in a node-adaptive way} to get the final propagated features.


\para{How to get deep ET?}
Lines of works are proposed to support large $D_t$, but none of them focuses on combining transformation outputs at different steps for each node. For example, JK-Net proposes a new transformation scheme for node embeddings that combines all previous node embeddings in the final layer.
Compared with MLP, ResNet and DenseNet directly access  the input signals in each previous layer, leading to the implicit deep supervision.
Motivated by them, other methods~\cite{DBLP:conf/icml/XuZJK21, li2019deepgcns, DBLP:conf/icml/Li0GK21, li2021deepgcns} adopt skip connections among GNN layers. For example, GCNII~\cite{chen2020simple} addresses small $D_t$ via initial residual connections and skip connections. Besides, as shown in Fig.~\ref{fig.res_acc}, both ResGCN and DenseGCN can support large $D_t$ with residual and dense connections, respectively.
Despite their effectiveness, the EP and ET operations are entangled in this type of methods, making them less scalable and efficient. 
We are motivated to ensure large $D_t$ while maintaining high efficiency and scalability. 
To this end, DGMLP also adopts \textbf{skip connections (i.e., residual connections in our model) among ET operations after obtaining the adaptively combined propagated features.}

\begin{table*}[]
\caption{Algorithm analysis. We denote $N$, $N_l$, $M$ and $d$ as the number of nodes, the number of labeled nodes, edges and feature dimensions respectively. P is the EP operation and T is the ET operation.
``SC'' means ``Skip Connection'', and $k$ refers to the number of sampled nodes in GraphSAGE.}
    \centering
    \resizebox{.85\linewidth}{!}{
    \begin{tabular}{l|c|c|c|c|c|c}
        \toprule
        \textbf{Algorithm} & \textbf{Convolution Type} & \textbf{Pattern} & \textbf{Disentangled}   & \textbf{Large $D_p$} & \textbf{Large $D_t$}& \textbf{Complexity}  \\
        \midrule
        GCN~\cite{kipf2016semi} & EPT & PT$\cdots$PT & $\times$  & $\times$& $\times$& $\mathcal{O}(D_pMd+D_tNd^2)$\\
        GraphSAGE~\cite{hamilton2017inductive} & EPT & PT$\cdots$PT & $\times$  & $\times$& $\times$&  $\mathcal{O}(k^{D_p}Nd^2)$\\
        \hline
        JK-Net~\cite{xu2018representation} & EPT-SC & PT$\cdots$PT & $\times$ &$\times$&\checkmark& $\mathcal{O}(D_pMd+D_tNd^2)$\\
        ResGCN~\cite{li2019deepgcns} & EPT-SC & PT$\cdots$PT & $\times$ &$\times$&\checkmark& $\mathcal{O}(D_pMd+D_tNd^2)$\\
        \hline
        APPNP~\cite{klicpera2018predict} & DTP & T$\cdots$TP$\cdots$P & \checkmark & \checkmark& $\times$& $\mathcal{O}(D_pMd+D_tNd^2)$\\
        AP-GCN~\cite{spinelli2020adaptive} & DTP & T$\cdots$TP$\cdots$P & \checkmark &\checkmark& $\times$& $\mathcal{O}(D_pMd+D_tNd^2)$\\
        DAGNN~\cite{liu2020towards} & DTP & T$\cdots$TP$\cdots$P & \checkmark &\checkmark& $\times$& $\mathcal{O}(D_pMd+D_tNd^2)$\\
        \hline
        SGC~\cite{wu2019simplifying} & DPT & P$\cdots$PT$\cdots$T & \checkmark &\checkmark& $\times$& $\mathcal{O}(D_tN_ld^2)$\\
        SIGN~\cite{frasca2020sign} & DPT & P$\cdots$PT$\cdots$T  & \checkmark &\checkmark& $\times$& $\mathcal{O}(D_tN_ld^2)$\\
        S$^2$GC~\cite{DBLP:journals/corr/abs-2004-11198} & DPT & P$\cdots$PT$\cdots$T  & \checkmark &\checkmark& $\times$& $\mathcal{O}(D_tN_ld^2)$\\
        GBP~\cite{chen2020scalable} & DPT & P$\cdots$PT$\cdots$T  & \checkmark &\checkmark& $\times$& $\mathcal{O}(D_tN_ld^2)$\\
        \hline
        DGMLP & DPT & P$\cdots$PT$\cdots$T & \checkmark  &\checkmark&\checkmark& $\mathcal{O}(D_tN_ld^2)$\\
        \bottomrule
    \end{tabular}}
    
    \label{algorithm analysis}
\end{table*}

\section{Proposed Solution}
Motivated by the above observations, we propose a scalable and flexible model termed Deep Graph Multi-Layer Perceptron (DGMLP), which contains a node-adaptive combination mechanism for increasing $D_p$ and the residual connections for large $D_t$.

\subsection{\sys Method}

\para{Deep propagation.}
Following the insight that different nodes require different optimal $D_p$, we design a node-adaptive combination mechanism that assigns varied weights to the propagated features at different steps for different nodes.
Similar to SGC, at a certain $D_p = k$, a smoothed feature vector $\mathbf{x}^{k}_v$ is collected from the features of the neighborhood $\mathcal{N}_v$ of node $v$:
\begin{equation}
\small
\begin{aligned}
 &\mathbf{x}^{k}_v =\sum_{u \in \mathcal{N}_v}\mathbf{x}_u^{k-1}/\sqrt{\tilde{d}_v\tilde{d}_u},
 \end{aligned}
\end{equation}
where $\mathbf{x}^0_v=\mathbf{x}_v$ is the original node feature. 
This process will be executed for $k$ times so that $x^k_v$ at propagation step $k$ could gather structural and feature information within $k$-hop neighborhoods.

We design node-adaptive weights for deeper propagation depth $D_p$.
More concretely, the \textit{propagation weight} $w_{v}(k)$ parameterized by node $v$ and propagation step $k$ is defined as the softmax output of
\{ $NSL_{v}(0)$, ${NSL_{v}(1)}$, $\cdots$, ${NSL_{v}(K)}$ \}:
\begin{equation}
\small
\label{iw}
w_{v}(k) = \frac{e^{{NSL_{v}(k)}/T}}{\sum \limits_{l=0}\limits^{K} e^{{NSL_{v}(l)}/T}},
\end{equation}
where $K$ is the maximal propagation step. Similar to Knowledge Distillation~\cite{hinton2015distilling, lan2018knowledge}, the temperature $T$ is adopted to soften or harden the probability distributions. Smaller $T$ will harden the distributions, and thus the model will focus more on the local graph information.
Besides, larger $w_{v}(k)$ means that $x^k_v$ maintains more original feature information of the node feature $x^0_v$, and are less likely to be over-smoothed. Therefore, for node $v$, the propagated features with larger $w_{v}(k)$ should contribute more to the final propagated features.

After obtaining the propagated features at different steps $\mathcal{M}_v=\{\mathbf{x}^{k}_v\ |\ 0\leq k \leq K\}$, we combine them into a single representation vector $m_v$:
\begin{equation}
\small
\mathbf{m}_v = \sum \limits_{l=0}\limits^{K} x^l_v w^l_v.
\end{equation}
By adaptively assigning different propagation weights for different nodes, we can simply increase $D_p$ on the graph level and get more powerful node embeddings at personalized smoothing levels for each node.

\para{Deep Transformation.}
Table~\ref{table.large_dt} motivates us that larger graphs require larger $D_t$ to get improved model performance. Besides, Fig.~\ref{fig.res_dense} shows both residual and dense connection can alleviate the \textit{model degradation} issue, thus helps to train a GNN with larger $D_t$. 
In this work, we select the residual connection for deep transformation. 
We refer to our transformation layer as the following format:
\begin{equation}
\small
\mathbf{h}^{(l+1)}_v = \sigma(\mathbf{h}^{(l)}_v\mathbf{W}^{(l)})+\mathbf{h}^{(l)}_v,
\end{equation}
where $\mathbf{W}^{(l)}$ is the learning parameter, $\mathbf{h}^{(0)}_v = \mathbf{m}_v$ is the original combined representation vector, and $\mathbf{h}^{(l)}_v$ is the transformed node embeddings of the $l$-th layer of the MLP with residual connections. 

\begin{table*}[tpb!]
\small
\centering
\caption{Overview of datasets and task types.}
\label{datasets}
\resizebox{.85\linewidth}{!}{
\begin{tabular}{cccccccc}
\toprule
\textbf{Dataset}&\textbf{\#Nodes}& \textbf{\#Features}&\textbf{\#Edges}&\textbf{\#Classes}&\textbf{\#Train/Val/Test}&\textbf{Description}\\
\midrule
Cora& 2,708 & 1,433 &5,429&7& 140/500/1,000 & citation network\\
Citeseer& 3,327 & 3,703&4,732&6& 120/500/1,000 & citation network\\
Pubmed& 19,717 & 500 &44,338&3& 60/500/1,000 & citation network\\
ogbn-arxiv& 169,343 & 128 & 1,166,243 & 40 &  91K/30K/47K 
& citation network\\
ogbn-products& 2,449,029 & 100 & 61,859,140 & 47 &  
196K/49K/2,204K
& citation network \\
ogbn-papers100M & 111,059,956 & 128 & 1,615,685,872 & 172 & 
1,207K/125K/214K & citation network\\
Industry & 1,000,000 & 64 & 1,434,382 & 253 & 5K/10K/30K&short-form video network\\
\bottomrule
\end{tabular}}
\end{table*}

\subsection{Comparison with Existing Methods}
\textbf{EPT-based GNNs.} GNNs with the convolution pattern of EPT assume that $D_p = D_t$. However, as discussed in Sec.~6.1, the optimal $D_p$ and $D_t$ are highly related to graph sparsity and graph size. For example, it is better to assign large $D_p$ and small $D_t$ for better performance on a small graph with sparse edges. 
Due to the \textit{model degradation} issue, current EPT-based GNNs usually have both small $D_p$ and $D_t$, owing to the inflexible restriction that $D_p = D_t$.
Besides, the EPT-based GNNs also face the problem of low scalability and low efficiency. For example, GCN has the high time complexity of $\mathcal{O}(D_pMd+D_tNd^2)$ for the need to repeatedly perform recursive neighborhood expansion at each training iteration to compute the hidden representations of each node.
This process is unscalable due to the high memory and computation costs on a single machine and high communication costs in distributed environments. Compared with EPT-based GNNs, DGMLP is more flexible in assigning different $D_p$ and $D_t$ and enjoys better efficiency, scalability, and lower memory requirement.

\noindent\textbf{EPT-SC-based GNNs.} Compared with EPT-based GNNs, the skip connection in EPT-SC-based GNNs helps to alleviate the \textit{model degradation} issue introduced by large $D_t$. However, Fig.~\ref{fig.res_acc} shows that the test accuracy degrades with deeper architecture due to the \textit{over-smoothing} issue introduced by large $D_p$. A large graph without the graph sparsity issue requires large $D_t$ and small $D_p$.
However, EPT-SC-based GNNs like ResGCN cannot satisfy such requirements since they simply restrict $D_p = D_t$.
Besides, EPT-SC-based GNNs still have the issues of low scalability, efficiency, and high memory requirement as EPT-based GNNs.
Compared with EPT-SC-based GNNs, DGMLP can improve the performance by enabling larger $D_p$ while maintaining high scalability, efficiency, and low memory requirement.

\noindent\textbf{DTP-based GNNs.} Compared with GNNs with the convolution pattern of EPT and EPT-SC, DTP-based GNNs disentangle the propagation and the transformation operation, thus can support flexible combinations of $D_p$ and $D_t$. 
As shown in Fig.~\ref{fig.smooth_nsl} and Fig.~\ref{fig.smooth_ak}, \textit{over-smoothing} may have less impact on the performance degradation compared with \textit{model degradation}, and the test accuracy will not decrease rapidly with the larger \textit{over-smoothing} level. 
Therefore, lots of DTP-based GNNs own large $D_p$ and small $D_t$ and claim that they can perform well with the deeper architecture. However, they ignore the transformation depth $D_t$ in their design. As shown in Fig.~\ref{fig:coupled}, the test accuracy degrades rapidly if we increase the transformation depth $D_t$ of DAGNN. Correspondingly, it is hard for DTP-based GNNs to support large graphs with limited transformation depth $D_t$. Besides, as they execute ET operations first, scalability, efficiency, and memory requirement issues still exist.

\noindent\textbf{DPT-based GNNs.} Compared with current DPT-based GNNs, the proposed DGMLP utilizes the same convolution pattern.
They all enjoy high scalability, efficiency, and low memory requirement as they can precompute the propagated features. The difference between DGMLP and other DPT-based GNNs is that DGMLP is the first method considering the smoothness from the node level.
Therefore, DGMLP can assign individual feature propagation weights to different nodes and then get better-smoothed node embeddings by considering the personalized area in the graph where each node resides. Another advantage of DGMLP is that it can support large transformation depth $D_t$ with the help of residual connections.

\begin{table*}[tpb!]
\caption{Test accuracy on the node classification task. ``OOM'' means ``out of memory''.}
\centering
{
\noindent
\renewcommand{\multirowsetup}{\centering}
\resizebox{0.9\linewidth}{!}{
\begin{tabular}{ccccccccc}
\toprule
\textbf{Type} & \textbf{Models} & \textbf{Cora} & \textbf{Citeseer} & \textbf{PubMed} & \textbf{Industry} & \textbf{ogbn-arxiv} & \textbf{ogbn-products} & \textbf{ogbn-papers100M}\\
\midrule
\multirowcell{2}{EPT}&
GCN& 81.8$\pm$0.5 & 70.8$\pm$0.5 &79.3$\pm$0.7 & 45.9$\pm$0.4 & 71.7$\pm$0.3 & OOM & OOM \\
&GraphSAGE& 79.2$\pm$0.6 & 71.6$\pm$0.5 & 77.4$\pm$0.5 & 45.7$\pm$0.6 &71.5$\pm$0.3 & \underline{78.3$\pm$0.2} & 64.8$\pm$0.4 \\
\midrule
\multirowcell{2}{EPT-SC}&
JK-Net& 81.8$\pm$0.5  & 70.7$\pm$0.7 & 78.8$\pm$0.7 & \underline{47.2$\pm$0.3} & 72.2$\pm$0.2 & OOM & OOM   \\
&ResGCN& 81.2$\pm$0.5  & 70.8$\pm$0.4 & 78.6$\pm$0.6 & 45.8$\pm$0.5 & \underline{72.6$\pm$0.4} & OOM & OOM    \\
\midrule
\multirowcell{3}{DTP}&
APPNP& 83.3$\pm$0.5 & 71.8$\pm$0.5 & 80.1$\pm$0.2 & 46.7$\pm$0.6 & 72.0$\pm$0.3 & OOM & OOM \\
&AP-GCN& 83.4$\pm$0.3& 71.3$\pm$0.5& 79.7$\pm$0.3 & 46.9$\pm$0.7 & 71.9$\pm$0.2 & OOM & OOM \\
&DAGNN & \underline{84.4$\pm$0.5} & \underline{73.3$\pm$0.6} & 80.5$\pm$0.5 & 47.1$\pm$0.6 & 72.1$\pm$0.3 & OOM & OOM \\
\midrule
\multirowcell{5}{DPT}&
SGC & 81.0$\pm$0.2 & 71.3$\pm$0.5 & 78.9$\pm$0.5 & 45.2$\pm$0.3 &71.2$\pm$0.3 & 75.9$\pm$0.2 & 63.2$\pm$0.2\\
&SIGN & 82.1$\pm$0.3 & 72.4$\pm$0.8 & 79.5$\pm$0.5 & 46.3$\pm$0.5 & 71.9$\pm$0.1 & 76.8$\pm$0.2 & 64.2$\pm$0.2\\
&S$^2$GC& 82.7$\pm$0.3 & 73.0$\pm$0.2 &79.9$\pm$0.3 & 46.6$\pm$0.6 & 71.8$\pm$0.3 & 77.1$\pm$0.1 & 64.7$\pm$0.3\\
&GBP& 83.9$\pm$0.7 & 72.9$\pm$0.5 &\underline{80.6$\pm$0.4} & 46.9$\pm$0.7 &72.2$\pm$0.2 & 77.7$\pm$0.2 & \underline{65.2$\pm$0.3}\\
&DGMLP & \textbf{84.6$\pm$0.6} & \textbf{73.4$\pm$0.5} & \textbf{81.2$\pm$0.6} & \textbf{47.6$\pm$0.7} & \textbf{72.8$\pm$0.2} & \textbf{78.5$\pm$0.2} & \textbf{65.7$\pm$0.2} \\
\bottomrule
\end{tabular}}}
\label{table.performance}
\end{table*}

\section{DGMLP Evaluation}
\subsection{Experimental Settings}
\noindent\textbf{Datasets.}
Three popular citation network datasets (i.e., Cora, Citeseer, and PubMed)~\cite{DBLP:journals/aim/SenNBGGE08} are used to benchmark the performance of DGMLP.
Besides, we also test our \sys on three large ogbn datasets: ogbn-arxiv, ogbn-products and ogbn-papers100M~\cite{hu2020ogb}.
It is worth noting that the ogbn-papers100M dataset contains more than 100 million nodes and 1 billion edges, which can be used as a criterion for whether a method is scalable.
At last, one real-world short-form video recommendation graph (Industry) from our industrial cooperative enterprise is also used in our experiments.
Table~\ref{datasets} presents an overview of these seven datasets.

\noindent\textbf{Parameters.}
For the three small citation networks, we adopt a simple logistic regression as the classifier.
The propagation depth $D_p$ is set to 20 for Cora and PubMed, and 15 for Citeseer.
The learning rate is set to 0.1, and the dropout rate is obtained from a search of range 0.1 to 0.5 with step 0.1.
Residual connections are not used on these three citation networks.

For three large ogbn datasets and Industry dataset, a 6-layer MLP with hidden size of 512 is used, and the propagation depth $D_p$ is respectively 20 and 12 for them.
Residual connections are used for these four large datasets.
Besides, the temperature $T$ is set to 1 as default if it is not specified, and the learning rate and the dropout rate are set to 0.001 and 0.5, respectively.
We run all the methods for 200 and 500 epochs on the citation networks and ogbn datasets, respectively.
Besides, we run all methods 10 times and report the mean values and the variances of different performance metrics.
Other hyperparameters are tuned with the toolkit OpenBox~\cite{li2021openbox} or follow the settings in their original paper.

\noindent\textbf{Baselines.}
We compare DGMLP with the following representative methods:
GCN~\cite{kipf2016semi}, GraphSAGE~\cite{hamilton2017inductive}, MLP, SGC~\cite{wu2019simplifying}, JK-Net~\cite{xu2018representation}, DAGNN~\cite{liu2020towards}, ResGCN~\cite{li2019deepgcns}, APPNP~\cite{klicpera2018predict}, AP-GCN~\cite{spinelli2020adaptive},  SIGN~\cite{frasca2020sign}, GBP~\cite{chen2020scalable}, and S$^2$GC~\cite{zhu2021simple}.
\subsection{End-to-End Comparison} 
The classification results on three citation networks are shown in Table~\ref{table.performance}. 
We observe that DGMLP outperforms all the compared baseline methods. 
Notably, DGMLP exceeds the current state-of-the-art model GBP by a margin of 0.6\% on the largest citation networks PubMed.
Compared with the EPT and EPT-SC type methods (i.e., GCN, JK-Net), the DTP-based GNNs (i.e., APPNP, AP-GCN) get better performance. 
This is due to the fact that the entanglement of transformation and propagation enables $D_p$ to go extremely deep, capturing more structural information.

We further evaluate DGMLP on the three popular OGB datasets, and the results are also summarized in Table~\ref{table.performance}.
The test accuracy of each method on the Industry dataset is also reported. 
It can be seen from Table~\ref{table.performance} that our proposed DGMLP constantly achieves the best performance across the four large datasets.
The improvement of DGMLP over baseline methods mainly relies on its support of large $D_p$ and $D_t$.

\subsection{Performance-Efficiency Analysis}

\begin{figure}
	\centering
	\includegraphics[width=.8\linewidth]{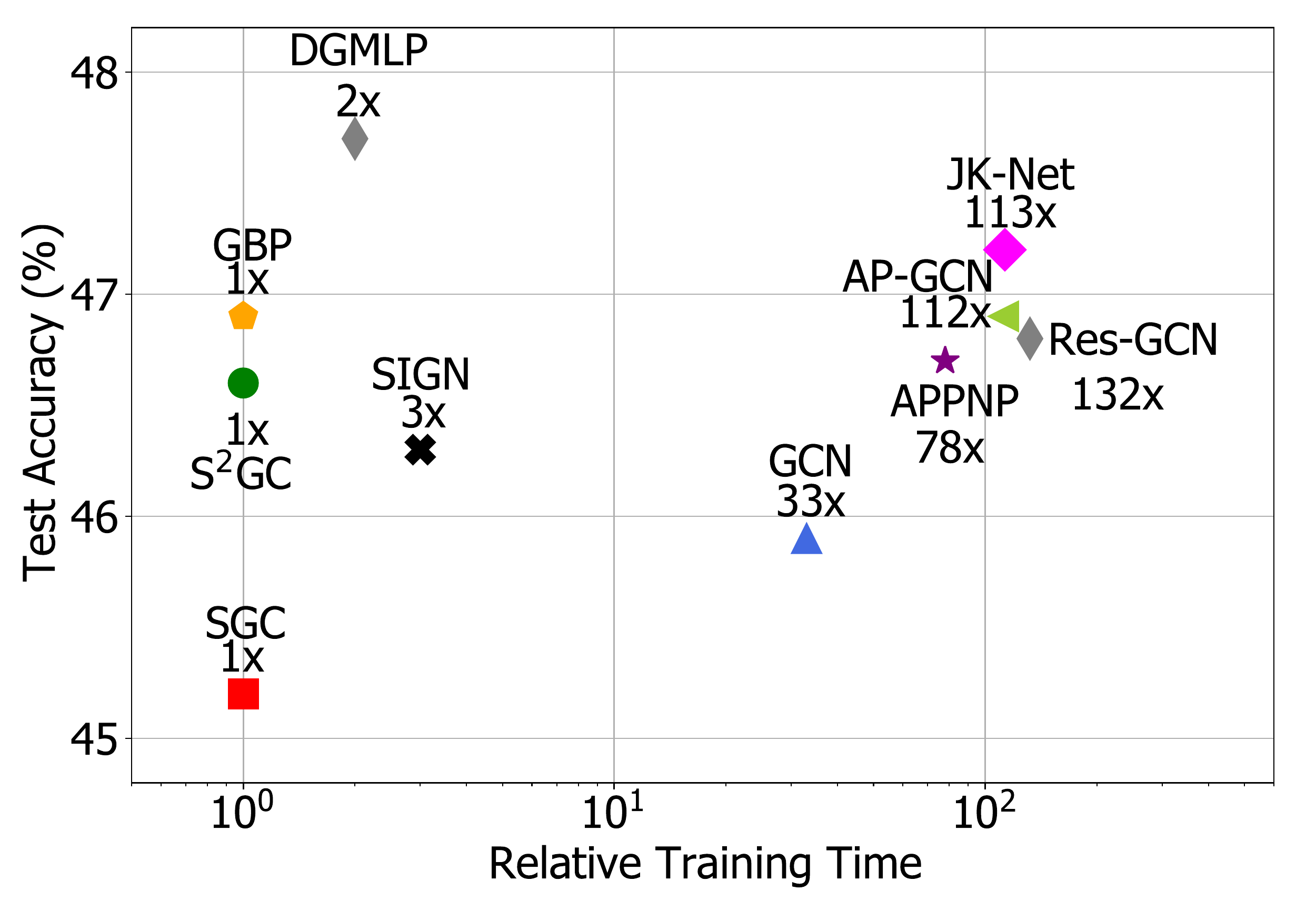}
	\caption{Performance along with training time on the Industry dataset.}
	\label{fig:efficiency}
\end{figure}

In this subsection, we evaluate the efficiency of each method in our industrial environment: the real-world Industry dataset.
Here, we precompute the smoothed features of each DPT-based GNN, and the time for preprocessing is also included in the training time.
Fig.~\ref{fig:efficiency} illustrates the results on the Industry dataset across representative baseline methods of each convolution pattern.

Compared with DPT-based GNNs, we observe that both EPT-based and DTP-based GNNs require a significantly larger training time.
For example, GCN takes 32 times longer than SGC to complete the training and the training time of AP-GCN is 112 times the one of SGC.
Due to the more complex preprocessing, our DGMLP takes twice the training time of SGC's.
However, the relatively time-consuming preprocessing brings significant performance gain to DGMLP, exceeding SGC by more than 2\% across three citation networks.
To sum up, our proposed DGMLP achieves better performance and maintains high efficiency.

\begin{figure}[tp!]
\centering  
\subfigure[efficiency]{
\label{fig.scala_effi}
\includegraphics[width=0.23\textwidth]{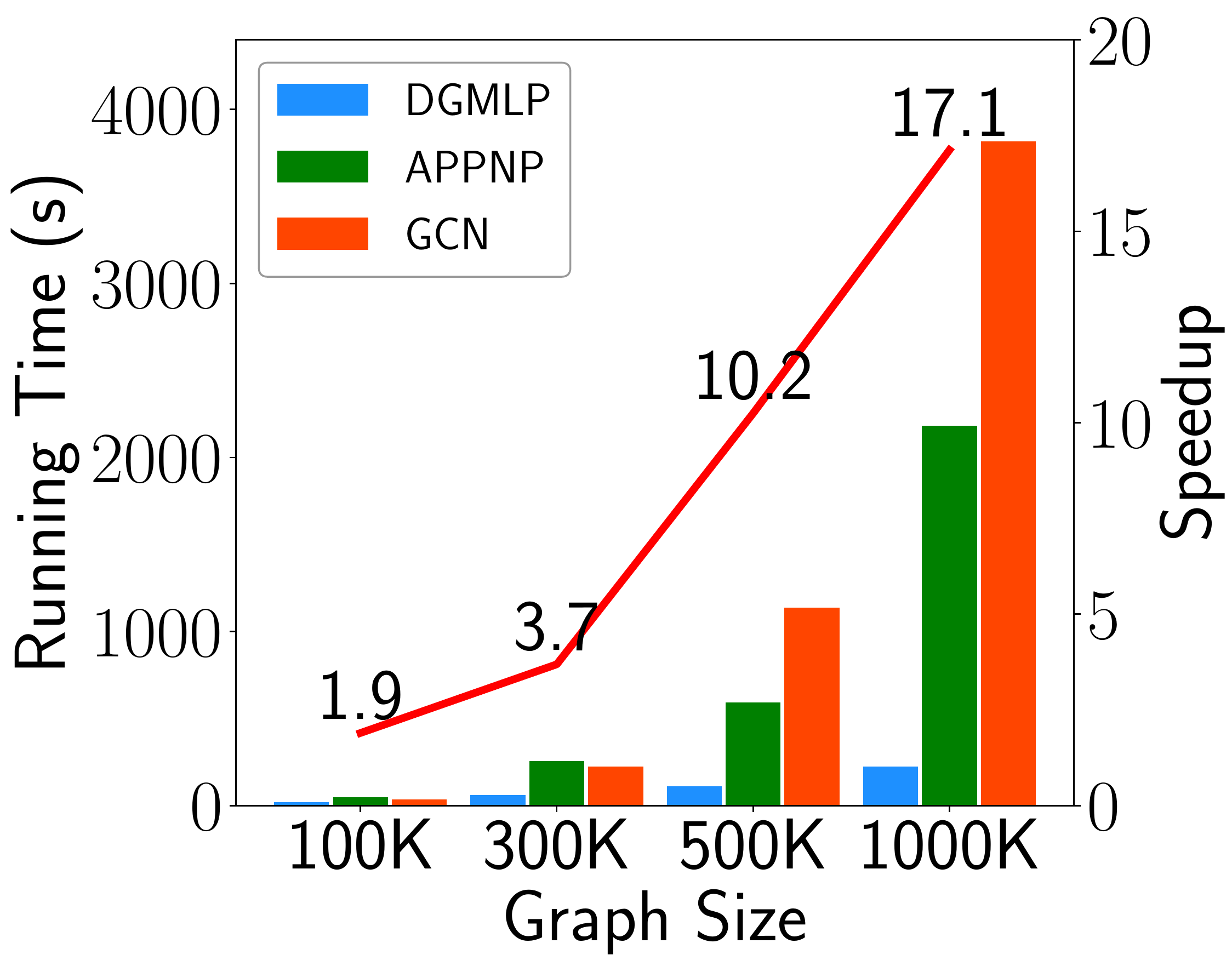}}\hspace{-1mm}
\subfigure[scalablility]{
\label{fig.scala_scala}
\includegraphics[width=0.23\textwidth]{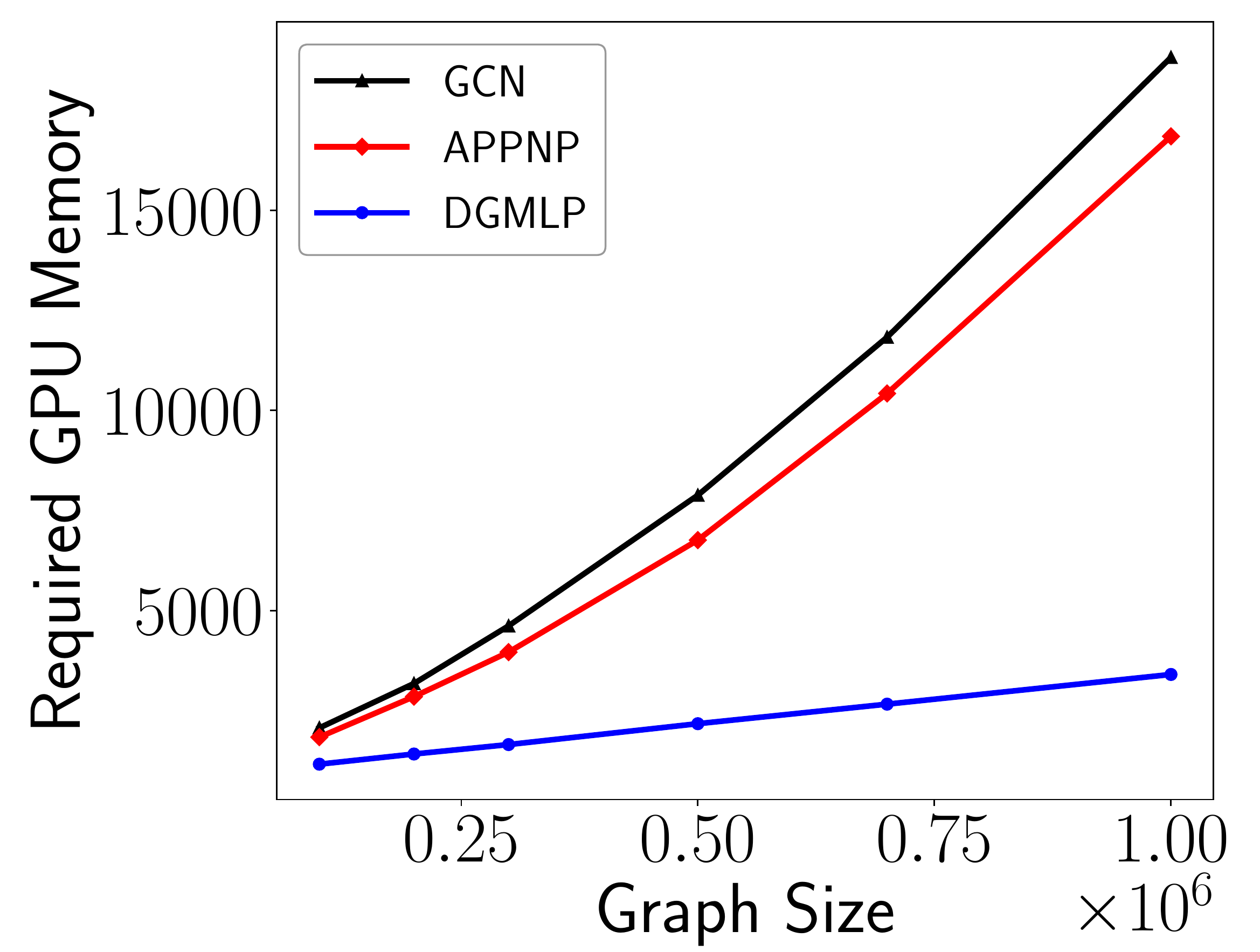}}\hspace{-1mm}
\caption{Running time and GPU memory requirement comparison on different sizes of graphs}
\label{fig.scalability}
\end{figure}

\subsection{Training Scalability}
To test the scalability of our proposed DGMLP, we use the Erdős-Rényi graph~\cite{erdos1960evolution} generator in the Python package NetworkX ~\cite{hagberg2008exploring} to generate artificial graphs of different sizes.
The node sizes of the generated artificial graphs vary from 0.1 million to 1 million, and the probability of an edge exists between two nodes is set to 0.0001.
As two representative EPT-based and DTP-based GNNs, GCN and APPNP are chosen as our baseline methods, and we run them for 200 epochs.
The total running time (including preprocessing time) and the tested GPU memory requirement are shown in Fig.~\ref{fig.scala_effi} and Fig.~\ref{fig.scala_scala}, respectively.
The running time speedup of DGMLP against GCN is also included in Fig.~\ref{fig.scala_effi}.

The experimental results in Fig.~\ref{fig.scala_effi} illustrate that DGMLP is highly efficient compared to GCN and APPNP.
It only takes DGMLP 223.4 seconds to finish the training on a large graph of size 1 million, which is less than the running time of both GCN and APPNP on the graph of size 0.3 million.
Fig.~\ref{fig.scala_scala} shows that the GPU memory requirement of DGMLP grows linearly with graph size.
However, the GPU memory requirements of GCN and APPNP both grow much quicker than DGMLP, exceeding 16GB when the graph size is 1 million while the memory requirement of DGMLP is just over 3GB at the same graph size. It indicates that our proposed DGMLP enjoys high scalability and efficiency at the same time.

\begin{figure}[tbp!]
\centering  
\subfigure[fix $D_t$ change $D_p$]{
\label{fig.dp_change}
\includegraphics[width=0.22\textwidth]{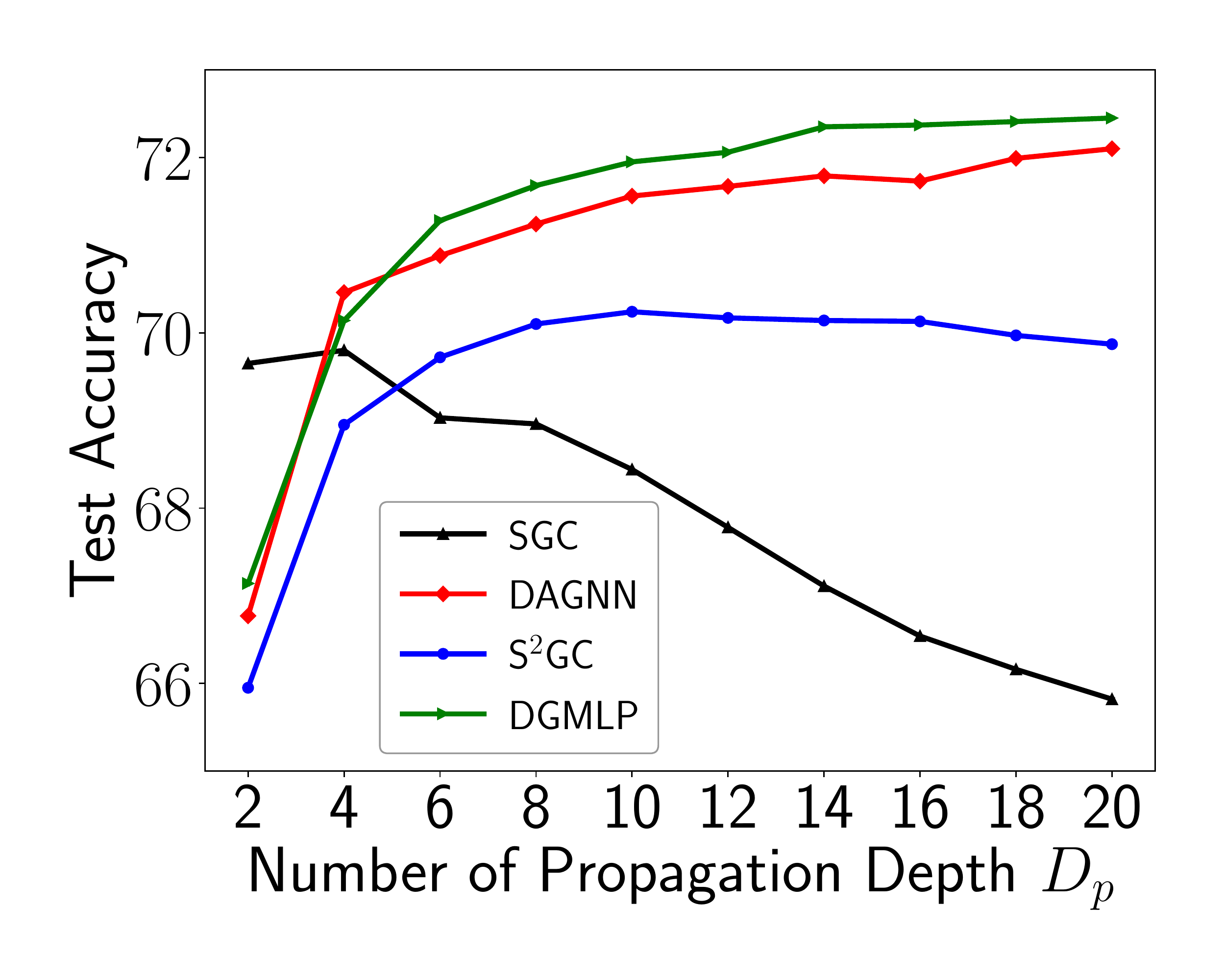}}\hspace{1mm}
\subfigure[fix $D_p$ change $D_t$]{
\label{fig.dt_change}
\includegraphics[width=0.22\textwidth]{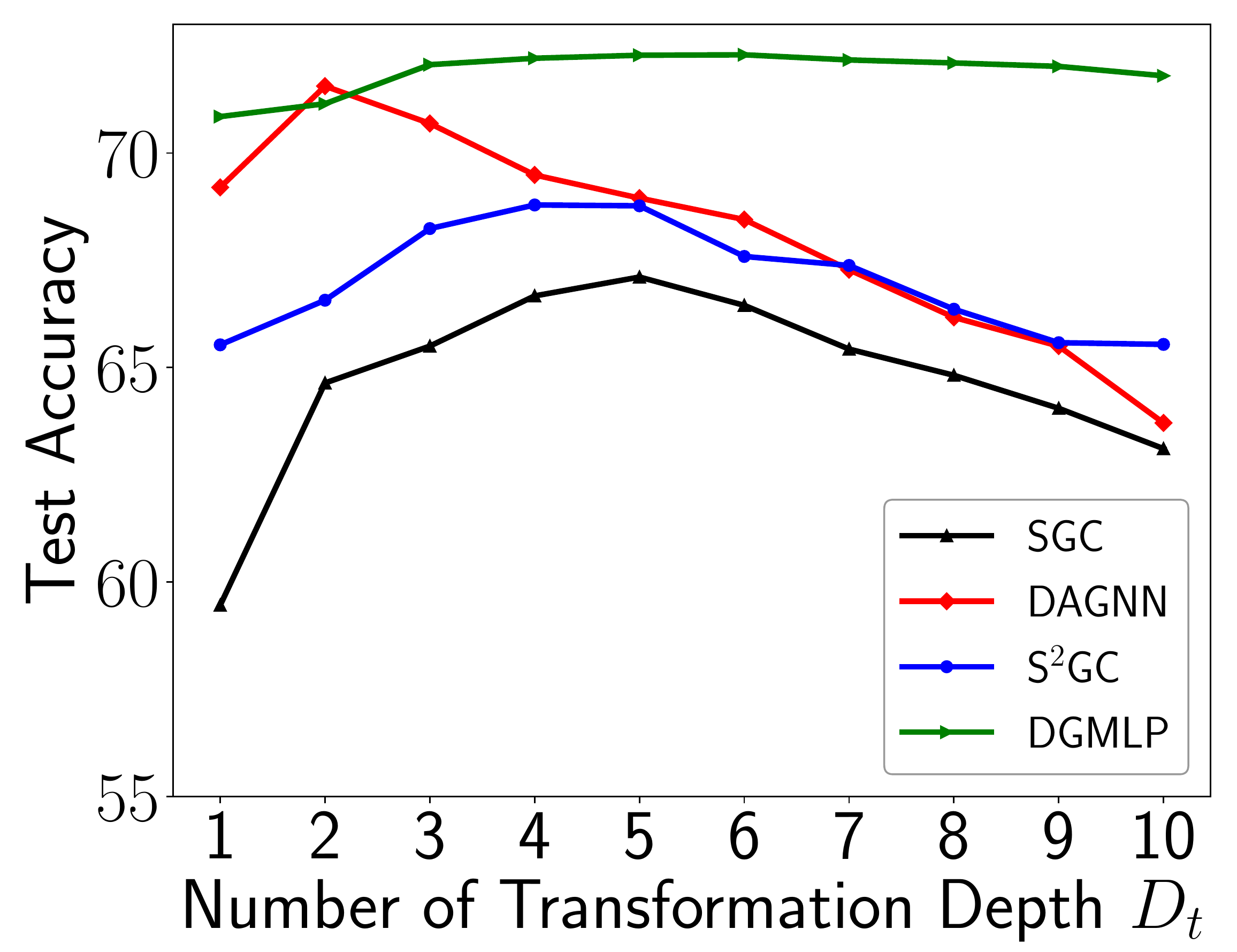}}\hspace{1mm}
\caption{Test accuracy with different $D_p$ or $D_t$.}
\label{fig.depth_ana}
\end{figure}

\begin{figure*}[tp!]
\centering  
\subfigure[Edge Sparsity]{
\includegraphics[width=0.31\textwidth]{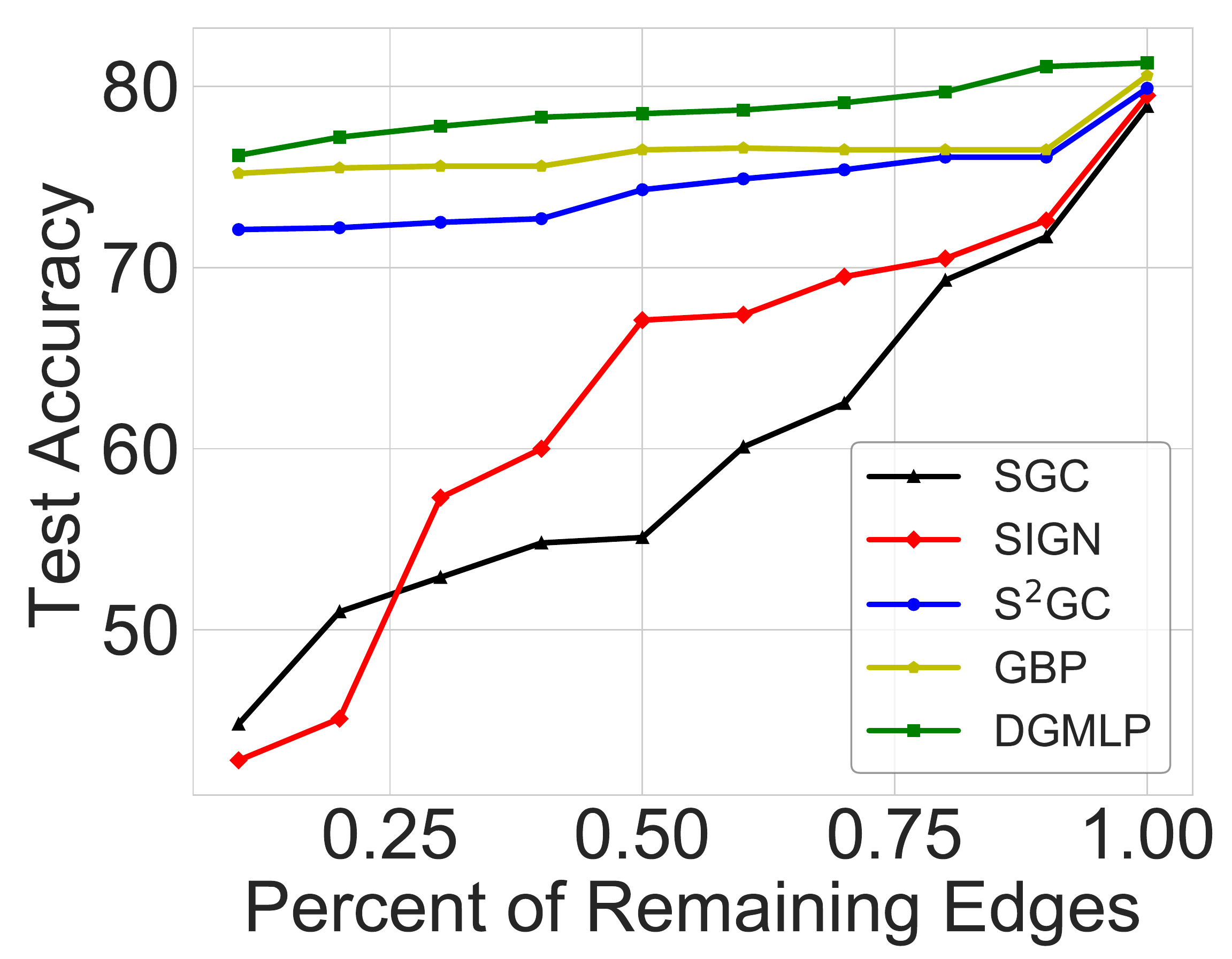}
\label{fig.sparse_edge}
}\hspace{-1mm}
\subfigure[Label Sparsity]{
\includegraphics[width=0.31\textwidth]{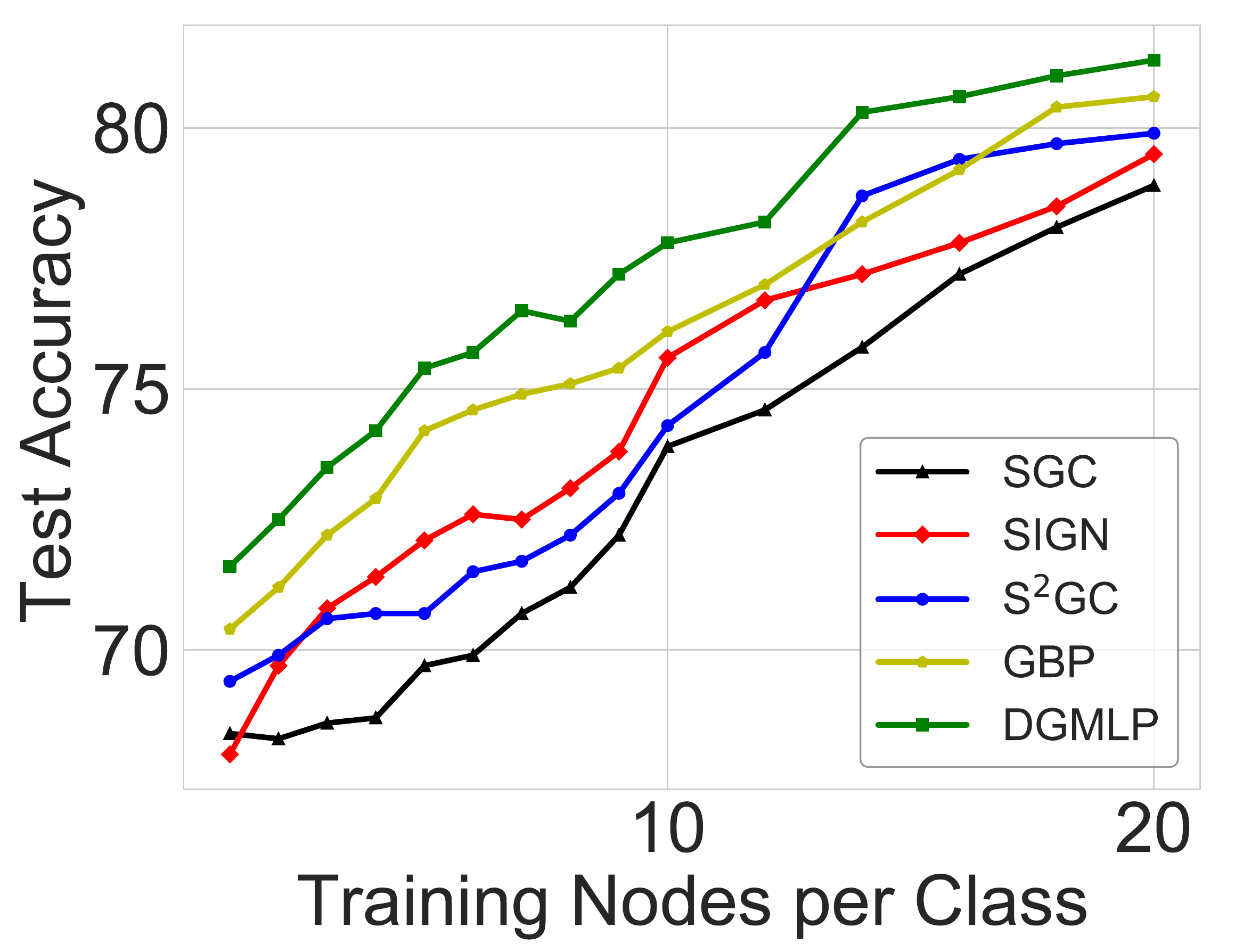}
\label{fig.sparse_label}
}\hspace{-1mm}
\subfigure[Feature Sparsity]{
\includegraphics[width=0.31\textwidth]{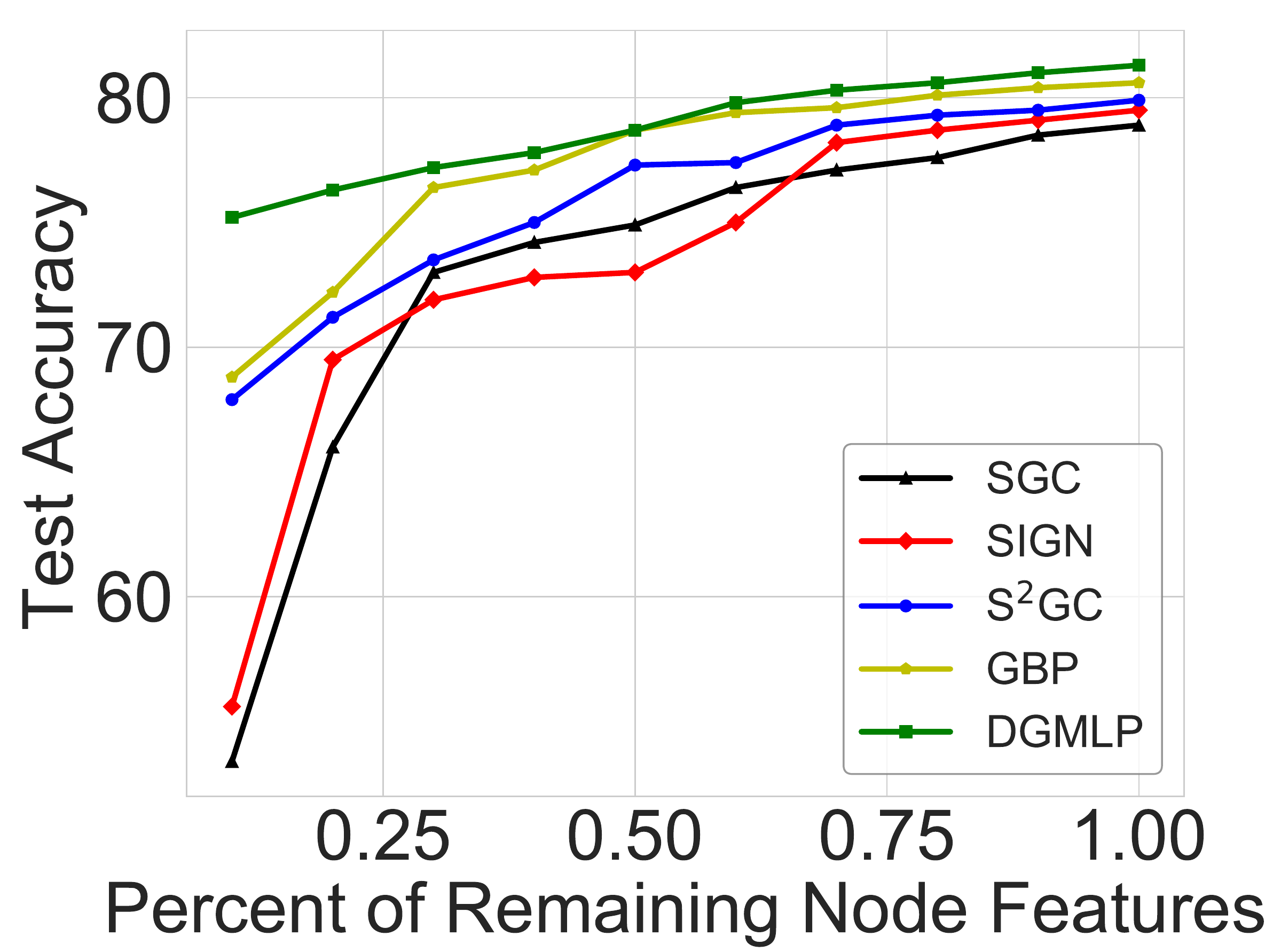}
\label{fig.sparse_feat}
}\hspace{-1mm}
\caption{Test accuracy on PubMed dataset under different levels of feature, edge and label sparsity.}
\label{fig.sparsity}
\end{figure*}

\begin{figure}[tpb!]
	\centering
	\includegraphics[width=.63\linewidth]{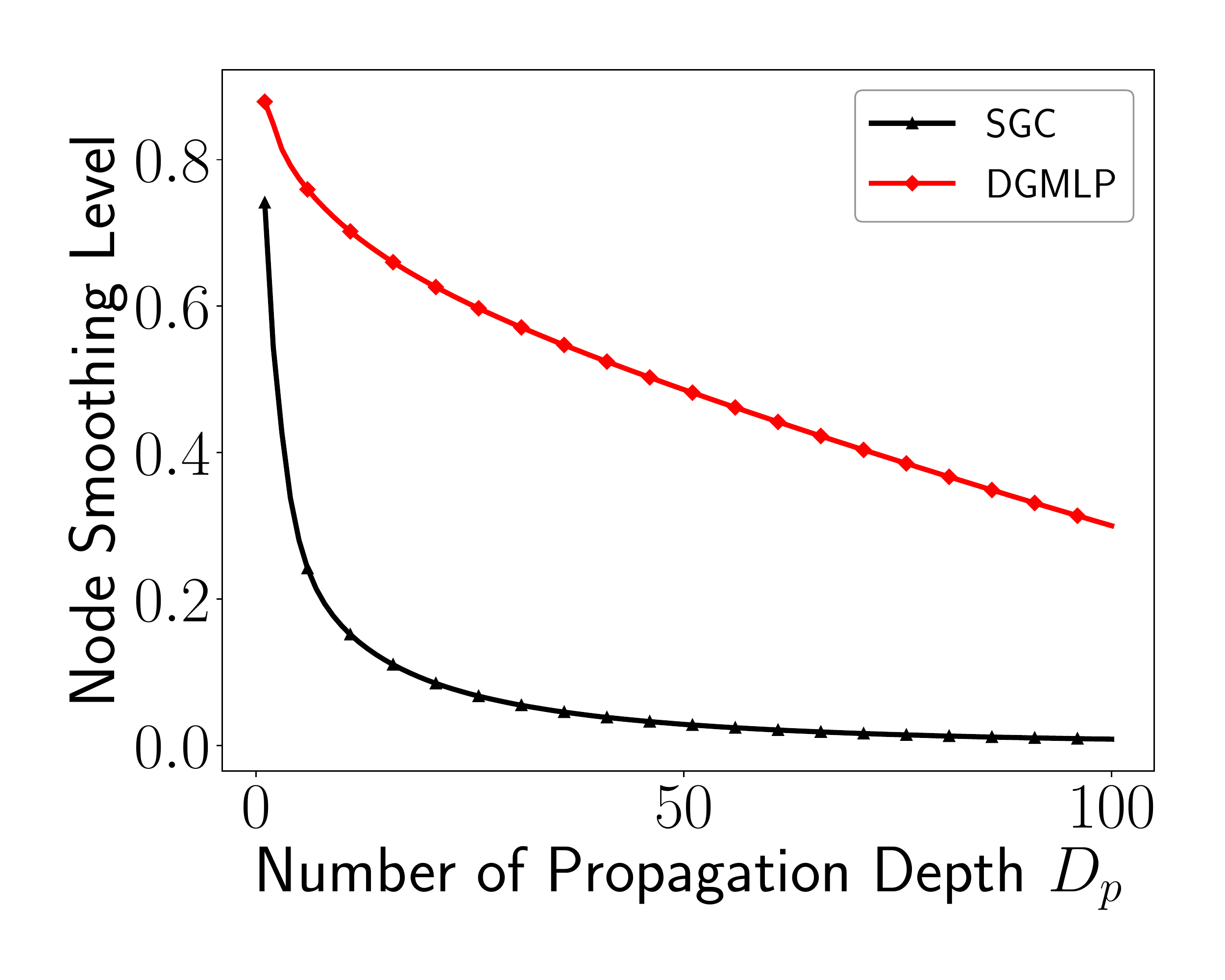}
	\caption{Graph Smoothing Level comparison between SGC and our proposed DGMLP.}
	\label{fig:interpret}
\end{figure}

\subsection{Analysis of model depth}
In this subsection, we conduct experiments to validate that our proposed DGMLP can go deep on both $D_p$ and $D_t$.
We choose DAGNN, SGC, and S$^2$GC as baseline methods.

Firstly, we fix the $D_t$ to 3 and increase the $D_p$ from 1 to 20 on the ogbn-arxiv dataset.
As seen from Fig.~\ref{fig.dp_change}, SGC cannot perform well when $D_p$ goes deep with the \textit{over-smoothing} issue.
Although the performance of DAGNN and S$^2$GC still maintains high when $D_p$ becomes large, our proposed DGMLP outperforms these two methods.
The superiority of our DGMLP over DAGNN and S$^2$GC lies in that we use a node-adaptive combination mechanism to directly control each node's smoothness.

Secondly, we fixed the $D_p$ to 10 and increase the $D_t$ from 1 to 10.
Fig.~\ref{fig.dt_change} shows that the performance of all the baseline methods, including SGC, DAGNN, and S$^2$GC degrades badly when $D_t$ becomes large.
It is because that these methods do not take \textit{model degradation} into consideration, which is precisely the main contributor to the performance drop when $D_t$ is large.
This property limits the expressive power of these baseline methods, resulting in relatively low performance when adopted on large graphs.
In the meantime, the performance of our proposed DGMLP still increases steadily or maintains even when $D_t$ is large.
To sum up, compared with other baselines, our method can consistently improve model performance with larger $D_p$ or $D_t$, which validates our insights and guidelines in Sec.~6.

\subsection{Influence of Graph Sparsity} 
To simulate extreme sparse situations in the real world, we design three independent settings on the PubMed dataset to test the performance of our proposed DGMLP when faced with edge sparsity, label sparsity, and feature sparsity, respectively.

\noindent\textbf{Edge Sparsity.} We randomly remove some edges in the original graph to strengthen the edge sparsity situation.
The edges removed from the original graph are fixed across all the methods under the same edge remaining rate.
From the results in Fig.~\ref{fig.sparse_edge}, it is clear to see that the performance of GBP, S$^2$GC, and our DGMLP is significantly better than SIGN and SGC.
Further, DGMLP always has higher test accuracy than GBP and S$^2$GC.
SIGN and SGC both have a flaw in that they can not effectively capture the deep structural information, which would become more prominent when edges are extremely sparse in the graph.

\noindent\textbf{Label Sparsity.} In this setting, we vary the training nodes per class from 1 to 20 and report the test accuracy of each method.
The experimental results in Fig.~\ref{fig.sparse_label} show that the test accuracy of all the compared methods increases as the number of training nodes per class becomes larger.
In the meantime, our DGMLP outperforms all the baselines throughout the experiment. With 11 training nodes per class, \sys achieves comparable performance with SGC trained under 20 training nodes per class.

\noindent\textbf{Feature Sparsity.} In a real-world situation, the feature of some nodes in the graph might be missing.
We follow the same experimental design in the edge sparsity setting but removing node features instead of edges.
The results in Fig.~\ref{fig.sparse_feat} illustrate that our proposed DGMLP has a great anti-interference ability when faced with feature sparsity as its performance drops only a little even there is only 10\% node features available.

\subsection{Interpretability} 
In this subsection, we empirically explain why our proposed DGMLP is more robust to the \textit{over-smoothing} issue.
Concretely, the baseline method SGC is used for comparison, and the temperature $T$ in our DGMLP is set to 0.2.
Besides, the Graph Smoothing Level ($GSL$) in Sec.~3.2.1 is adopted to evaluate the graph smoothness.

The results in Fig.~\ref{fig:interpret} show that the $GSL$ of both SGC and DGMLP decreases as $D_p$ increases.
However, the descending speed of SGC is much quicker than DGMLP, especially when $D_p$ changes from 1 to 5. Moreover, the $GSL$ of DGMLP at $D_p=100$ is even larger than the one of SGC at $D_p=5$.
Fig.~\ref{fig:interpret} strongly illustrates that our proposed DGMLP is more robust to the \textit{over-smoothing} issue introduced by large $D_p$, and the performance results in Fig.~\ref{fig.dp_change} further shows that DGMLP can take advantage of this property to gain more beneficial deep structural information for prediction.

\section{Conclusion}
In this paper, we make an experimental evaluation of current GNNs and find the true limitations of GNN depth: the \textit{over-smoothing} introduced by large propagation depth and the \textit{model degradation} introduced by large transformation depth. Besides, experiments show that sparse graphs require larger propagation depth, and it is better to assign larger transformation depth on large graphs. Based on the above analysis, we present Deep Graph Multi-Layer Perceptron (DGMLP), a flexible and deep graph model that can simultaneously support large propagation and transformation depth. Extensive experiments on seven real-world graph datasets demonstrate the high accuracy, scalability, efficiency of DGMLP against the state-of-the-art GNNs.


\bibliographystyle{ACM-Reference-Format}
\balance
\bibliography{reference}

\end{document}